\documentclass[letterpaper]{article} 
\usepackage{aaai2026}  
\usepackage{times}  
\usepackage{helvet}  
\usepackage{courier}  
\usepackage[hyphens]{url}  
\usepackage{graphicx} 
\usepackage{natbib}  
\usepackage{caption} 
\usepackage{algorithm}
\usepackage{algorithmic}
\usepackage{multirow}
\usepackage{algorithm}
\usepackage{algorithmic}
\usepackage{amssymb}
\usepackage{amsmath}
\usepackage{graphicx}
\usepackage{bm}
\usepackage{booktabs}
\usepackage{cleveref}
\usepackage{tcolorbox}
\usepackage{subcaption}

\usepackage{newfloat}
\usepackage{listings}
\DeclareCaptionStyle{ruled}{labelfont=normalfont,labelsep=colon,strut=off} 
\floatstyle{ruled}
\newfloat{listing}{tb}{lst}{}
\floatname{listing}{Listing}
\title{Ellipsoid-Based Decision Boundaries for Open Intent Classification}
\author {
    Yuetian Zou\textsuperscript{\rm 1}\equalcontrib,
    Hanlei Zhang\textsuperscript{\rm 1}\equalcontrib,
    Hua Xu\textsuperscript{\rm 1}\thanks{This is the corresponding author.},
    Songze Li\textsuperscript{\rm 2},
    Long Xiao\textsuperscript{\rm 2}
}
\affiliations {
    \textsuperscript{\rm 1}Tsinghua University\\
    \textsuperscript{\rm 2}Hebei University of Science and Technology\\
    zouyt21@mails.tsinghua.edu.cn, zhang-hl20@mails.tsinghua.edu.cn, xuhua@mail.tsinghua.edu.cn
}

\begin{document}

\maketitle

\begin{abstract}
Textual open intent classification is crucial for real-world dialogue systems, enabling robust detection of unknown user intents without prior knowledge and contributing to the robustness of the system. While adaptive decision boundary methods have shown great potential by eliminating manual threshold tuning, existing approaches assume isotropic distributions of known classes, restricting boundaries to balls and overlooking distributional variance along different directions. To address this limitation, we propose EliDecide, a novel method that learns ellipsoid decision boundaries with varying scales along different feature directions. First, we employ supervised contrastive learning to obtain a discriminative feature space for known samples. Second, we apply learnable matrices to parameterize ellipsoids as the boundaries of each known class, offering greater flexibility than spherical boundaries defined solely by centers and radii. Third, we optimize the boundaries via a novelly designed dual loss function that balances empirical and open-space risks: expanding boundaries to cover known samples while contracting them against synthesized pseudo-open samples. Our method achieves state-of-the-art performance on multiple text intent benchmarks and further on a question classification dataset. The flexibility of the ellipsoids demonstrates superior open intent detection capability and strong potential for generalization to more text classification tasks in diverse complex open-world scenarios. 
\end{abstract}

\begin{links}
    \link{Code}{https://github.com/thuiar/textoir}
\end{links}


\section{Introduction}

\begin{figure}
    \centering
    \includegraphics[width=\linewidth]{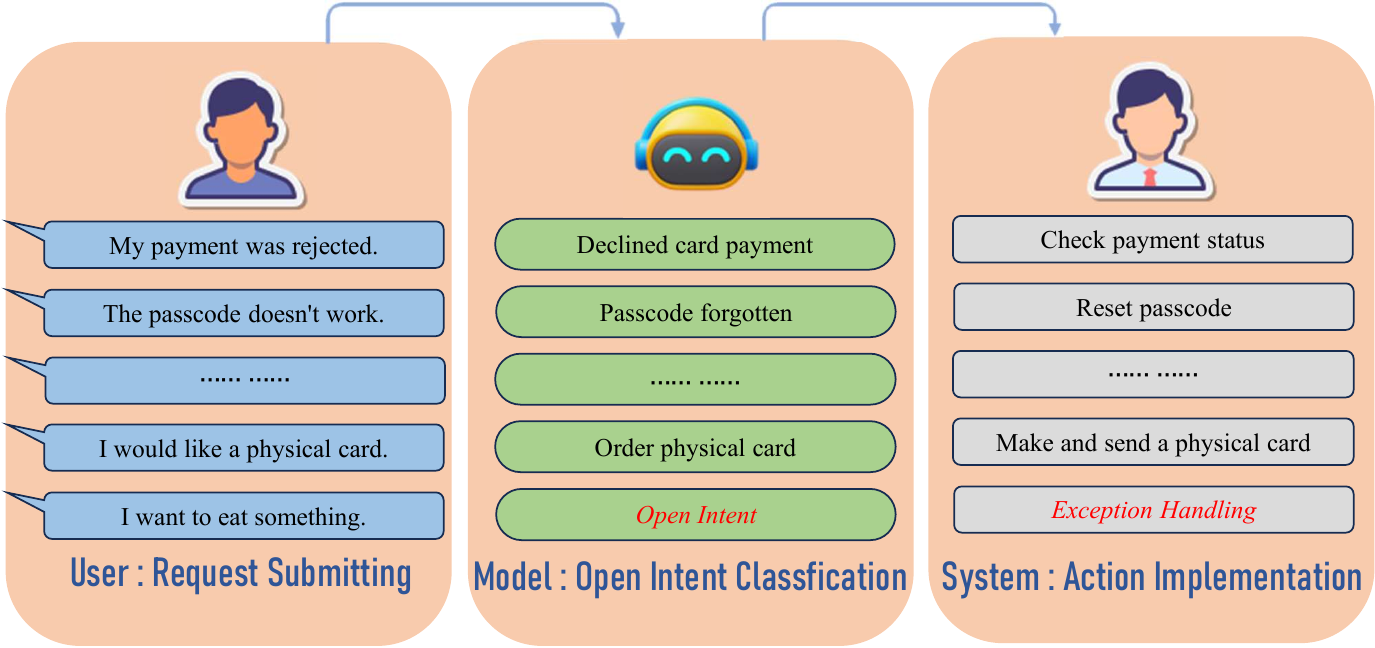}
    \caption{Open Intent Classification in Banking: Model classifies user requests into known intents (subsequently triggering corresponding services), or detects unknown (\textit{open}) intents (handled as exceptions). }
    \label{fig:intro1}
\end{figure}

Open world classification is critical for robust real-world systems, as it enables accurate recognition of known classes while simultaneously rejecting anomalies to reduce open-space risks~\citep{yang2024generalized}. This capability is essential across diverse domains, including autonomous driving~\citep{geiger2012we} and medical image analysis~\citep{zimmerer2022mood}. In the realm of natural language understanding, dialogue systems specifically require the ability to recognize the known user intents while detecting unknown ones, which is a significant capability for system robustness and continuous improvement~\citep{lin2019deep}, as shown in Figure~\ref{fig:intro1}. Consequently, to develop robust dialogue systems, extensive research~\citep{lin2019deep,larson2019evaluation,zhang2021deep,zhang2023learning,zhou2022knn,zhou2023two} has focused on open world classification for user intents, known as \textit{Open Intent Classification}. In this work, we formulate open world classification as a $(K+1)$-way classification task comprising $K$ known classes and a single unknown (\textit{open}) class. This formulation aligns with the established paradigm of Open Set Recognition (OSR)~\citep{scheirer2012toward}.

\begin{figure}
    \centering
    \includegraphics[width=0.8\linewidth]{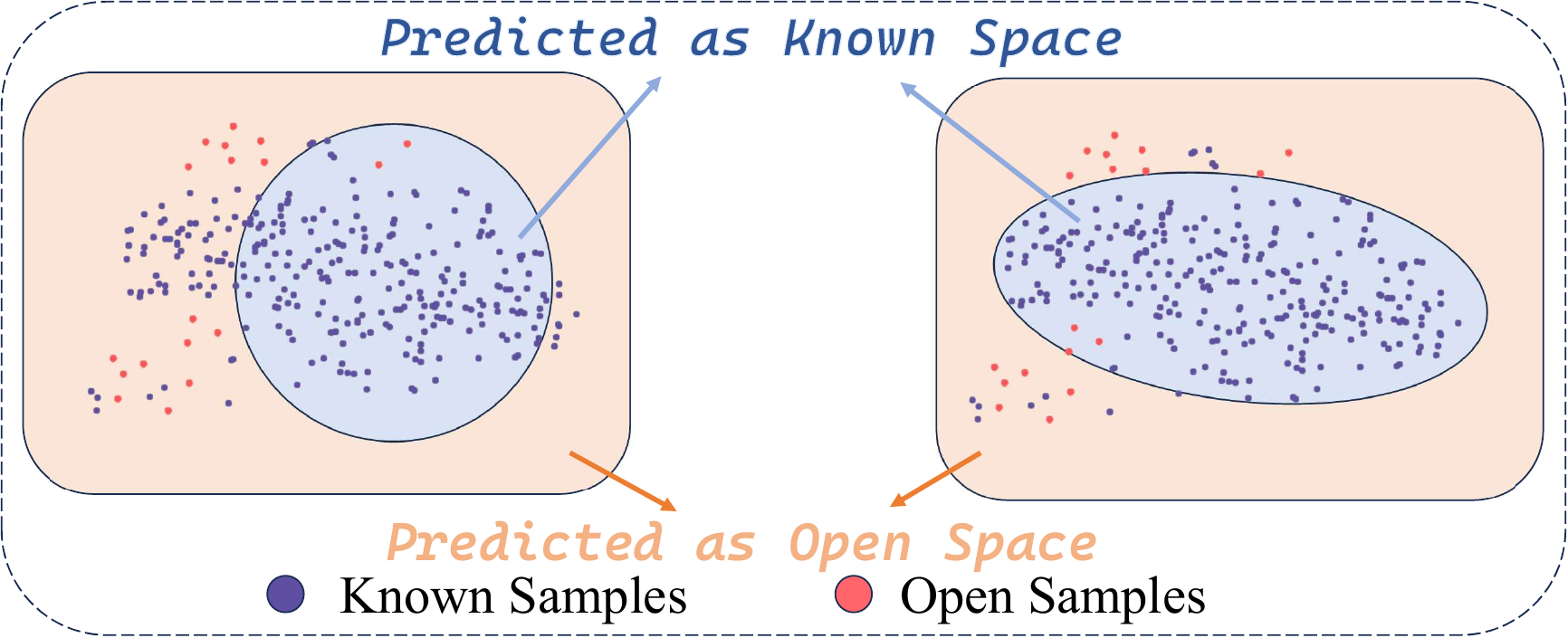}
    \caption{An example of a known class's anisotropic feature distribution. The ball boundary excludes a significant portion of known samples as the cost of avoiding open ones inside, while the ellipsoid includes most of the known samples without many open ones inside due to geometric flexibility.}
    \label{fig:intro2}
\end{figure}

While the early open world classification methods~\cite{bendale2016towards, hendrycks2016baseline, liang2017enhancing} relied on class probability distributions, recent approaches utilize deep representations from pretrained models, which primarily fall into two categories. Scoring-based approaches~\cite{zhou2022knn,zhou2023two,Zhang_Bai_Li_2023,JOSC,gautam-etal-2024-class} construct discriminative feature spaces and identify unknown samples through statistic-based scoring functions like local density measures, but they require manual threshold tuning, limiting their practical use. In contrast, boundary-based methods~\cite{zhang2021deep,zhang2023learning,Liu_Li_Mu_Yang_Xu_Wang_2023,chen_2024,Li_Ouyang_Pan_Zhang_Zhao_Xia_Yang_Wang_Li_2025} avoid thresholds by learning adaptive decision boundaries. However, current boundary-based approaches typically assume isotropic feature distributions, ignoring the directional variance inherent in real-world data. As illustrated in Figure~\ref{fig:intro2}, ellipsoid boundaries offer a more flexible representation than spherical ones, yielding lower open-space and empirical risks while producing more concise closed regions.

In this paper, we propose EliDecide, an ellipsoid-based decision boundary learning method for open-world classification. First, supervised contrastive learning is applied to a pre-trained model to obtain discriminative representations that serve as the basis for adaptive boundary construction. In the feature representation space, we construct ellipsoid-shaped boundaries for each known class and regard the outside of all ellipsoids as the open space. Each ellipsoid is parameterized by a learnable nonsingular matrix encoding both the directions and the lengths of the ellipsoid's axes. We optimize the boundaries using a dual loss mechanism. A positive loss expands the boundary when known samples (used as positive samples) fall outside it to encourage it to involve more known samples, while a negative loss contracts the boundary when pseudo-open samples (used as negative examples) lie too close or within it to prevent it from over-expanding and involving unknown samples. This integrated strategy enables EliDecide to adapt decision boundaries to the inherent directional variance in real-world data, representing the distributions of known samples flexibly and precisly and achieving robust open-world classification. 

We extensively evaluate EliDecide on benchmark intent datasets and demonstrate its superior performance and robustness compared to state-of-the-art methods. Beyond intent, experiments on a question classification dataset show the generalization potential of our method in more diverse and complex open-world classification tasks. Additional experiments including comparisons among boundaries of different shapes further validate the advantage of the ellipsoid formulation, highlighting its effectiveness and robustness across different open-world scenarios.

We summarize our contributions as follows: 

\begin{itemize}
    \item To overcome the limitation of spherical boundaries in existing methods, we introduce the first ellipsoid boundary method for open world classification. We propose an effective parameterization that represents an ellipsoid via a learnable matrix capturing both axis directions and lengths. Novel and effective dual loss mechanism is designed to optimize the parameter matrices of boundaries.
    \item We evaluate EliDecide on two benchmark intent datasets as well as a question classification dataset, achieving a new state-of-the-art performance across all tasks. This demonstrates its robust effectiveness and strong generalization capabilities.
    \item Comprehensive additional experiments and analysis including geometric interpretation, ablation studies and performance comparisons with Large Language Models (LLMs) validate the theoretical foundations and practical superiority of our methodology.
\end{itemize} 

\section{Related Works}\label{sec:related_work}

Based on the primary detection criterion, existing methods fall into two categories: probability-based methods and deep representation-based methods. 

\subsection{Probability-Based Methods}

Early-stage methods detect open samples using class probability distributions. For example, OpenMax~\cite{bendale2016towards} modifies softmax networks by replacing the penultimate layer with activation vectors to obtain the probabilities of $(K+1)$-way classification task. Softmax-MSP~\citep{hendrycks2016baseline} demonstrates that softmax probabilities can serve as a baseline for distinguishing between known and unknown data, while ODIN~\citep{liang2017enhancing} improves this approach by applying temperature scaling and input perturbations. However, these methods rely solely on output probabilities and fail to leverage the deep semantic features.

\subsection{Deep Representation-Based Methods}

Recent approaches overcome the limitations of softmax-based classifiers by exploiting deep semantic features extracted by pre-trained models. These methods can be further categorized as \textit{scoring-based methods} and \textit{adaptive boundary-based methods}. 

\textbf{Scoring-based methods} learn discriminative feature representations to calculate confidence scores and discriminate between known and open samples by comparing the scores against a threshold. For example, DOC~\cite{shu2017doc} employs a one-vs-rest sigmoid layer to detect open samples. \citet{lin2019deep} first explores open intent detection using bidirectional LSTMs with local outlier factor (LOF) techniques. KNNCL~\cite{zhou2022knn} uses contrastive learning to improve the discriminability of semantic features. DE~\cite{zhou2023two} addresses deep overconfidence by voting based-early exiting. SCOOS~\cite{gautam-etal-2024-class} leverages the embeddings of the intent labels as semantic cues. Both KNNCL and DE adopt the density-based scoring functions like LOF for detecting open samples while SCOOS uses statistic-based thresholds. However, manual threshold tuning hinders their practical deployment.

\begin{figure*}
    \centering
    \includegraphics[width=0.85\textwidth]{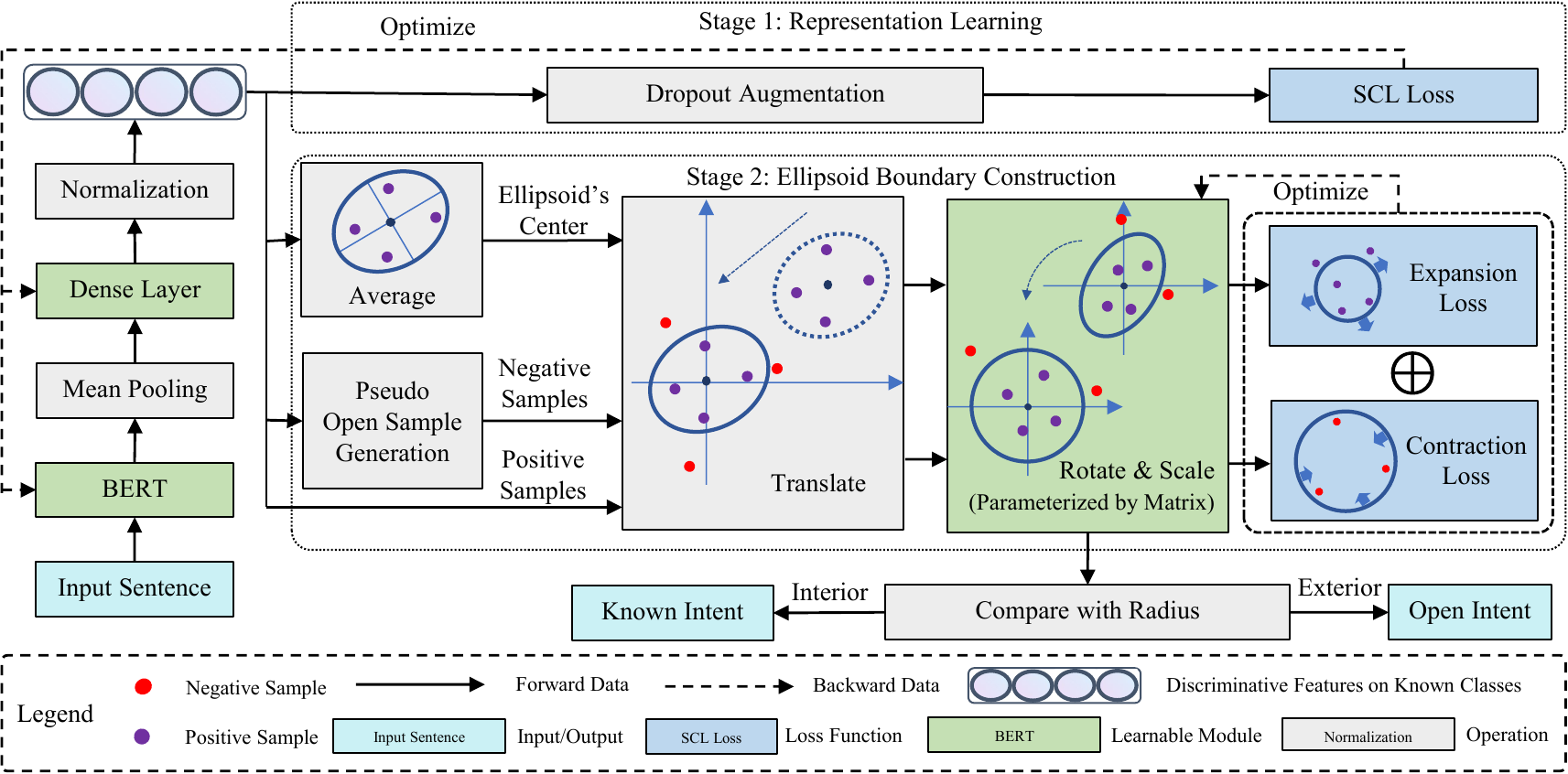}
    \caption{The structure of our method. The first stage is to learn discriminative representations by SCL and then learn ellipsoid-based boundaries with expansion and contraction losses by using both positive and pseudo-negative samples. }
    \label{fig:method}
\end{figure*}

\textbf{Adaptive boundary-based methods} dynamically adjust decision boundaries instead of relying on manually tuned thresholds. \citet{zhang2021deep} further formulates the open intent classification task and proposes ball-like adaptive decision boundaries to reduce open-space risk. On this basis, to refine the feature representation for the boundaries, DA-ADB~\cite{zhang2023learning} introduces Distance-Aware Representation Learning and TCAB~\cite{chen_2024} applies a triplet-contrastive learning strategy. CLAB~\cite{Liu_Li_Mu_Yang_Xu_Wang_2023} designs a novel loss function with expanding and shrinking operations to learn more suitable boundaries. MOGB~\citep{Li_Ouyang_Pan_Zhang_Zhao_Xia_Yang_Wang_Li_2025} establishes multiple spherical boundaries for individual classes. Different from the existing methods with spherical boundaries, our proposed method introduces a more flexible and adaptive ellipsoidal boundaries that provide discriminative criteria for this challenging task.

\section{Proposed Method}
Our proposed method aims to build an ellipsoid-shaped boundary to describe the distribution of each known class in the feature space. The complementary space to all the ellipsoids will be regarded as the open space. Our method can be separated into two stages as shown in Figure~\ref{fig:method}. In the first stage, we apply supervised contrastive learning~\cite{khosla2020supervised} to learn discriminative representations for boundary learning. Then, we construct ellipsoid-based boundaries and optimize them with novelly designed dual optimization objectives by using both positive and pseudo-negative samples.

\subsection{Representation Learning}\label{sec:finetune}

Following~\cite{zhang2021deep,zhou2022knn,zhou2023two}, we apply pre-trained BERT model~\cite{Devlin_Chang_Lee_Toutanova_2019} as the backbone for feature extraction and use the average representations of the last layer of the encoder to yield the sentence-level representation:
\begin{equation}
    \label{eq:feature}
        x_i=\frac{1}{N+1}(\text{[CLS]}+\sum_{j=1}^N T_j),
        z_i=\frac{Wx_i+b}{||Wx_i+b||_2},
\end{equation}
where $N$ is the number of tokens in the sentence, $\text{[CLS]}$ and $\{T_i\}_{i=1}^N\in \mathbb{R}^H$ are token-level features, $H$ is the hidden size. As suggested in~\cite{zhang2023learning}, we add a linear layer with the weight matrix $W\in\mathbb{R}^{H\times n}$ and bias vector $b\in \mathbb{R}^{n}$ to enhance the vanilla representation $x_i$, where $n$ is the dimension of the representation. Normalization is applied to ensure compactness of the representation space and harmonize the scales of the distributions among classes, yielding the final representation $z_i\in \mathbb{R}^n$. 

With the motivation of pushing intra-class compact and inter-class separation, we apply supervised contrastive learning~\cite{khosla2020supervised} as below:
\begin{equation}
\mathcal{L}_{\textrm{con}} =-\sum_{i\in I} \frac{1}{|P_i|}\sum_{p\in P_i} \log\frac{\exp(z_i\cdot z_p/\tau)}{\sum_{a\in A_i}{\exp(z_i\cdot z_a/\tau)}},
\end{equation}
where $I$ are the sample indices in a mini-batch, $A_i=I\setminus\{i\}$ and $P_i=\{p\in A(i)|y_p=y_i\}$. Here we apply $\textit{dropout augmentation}$~\cite{gao2021simcse} to obtain the positive augmentation of each sample. By minimizing $\mathcal{L}_{\text{con}}$, the representations of samples with the same label are encouraged to converge, whereas those of samples with distinct labels are induced to diverge. 

\subsection{Ellipsoid Boundary Construction}\label{sec:boundary}

For each known class \(k\), we denote the region inside its ellipsoid boundary as \(E_k \subset \mathbb{R}^n\). To construct this boundary, we first compute the centroid of the ellipsoid and then learn a matrix that encodes its \textit{shape}, including both its size and orientation. Simultaneously, we construct an affine transformation that maps the ellipsoid to a ball, which is useful for determining whether a point lies within the ellipsoid and for calculating the proposed loss function.

\subsubsection{Centroid Computation}

For each known class \(k\), we compute the ellipsoid centroid \(c_k\) as the mean of the representations of samples belonging to class \(k\):
\begin{equation}
\label{eq:center}
    c_k = \frac{1}{|S_k|}\sum_{z_i \in S_k} z_i,
\end{equation}
where \(S_k = \{ z_i \mid y_i = k \}\) denotes the representation set of class \(k\). 

\subsubsection{Shape Parameterization via Learnable Matrix}

In this section we parameterize an ellipsoid $\widetilde{E_k}$ with the center at the origin, namely a \textit{centralized ellipsoid}, as a pattern of the shape (size and orientation) of $E_k$. Then $E_k$ is defined by a simple translation:
\begin{equation}
\label{eq:centralized_ellipsoid}
    E_k = \widetilde{E_k} + c_k := \{ \hat{x} \in \mathbb{R}^n \mid \hat{x} = x + c_k,\ x \in \widetilde{E_k} \}.
\end{equation}
A centralized ellipsoid is commonly characterized by \(n\) orthogonal unit vectors \(\{d_i\}_{i=1}^n\) specifying its axis directions and corresponding positive axis lengths \(\{a_i\}_{i=1}^n \subset \mathbb{R}^+\). However, directly optimizing the orthogonal vectors \(\{d_i\}\) using gradient descent is intractable due to the rigid orthogonality constraints. To overcome this, we propose to learn a matrix \(A_k \in \mathbb{R}^{n \times n}\) that implicitly encodes both the axis directions and scales and define the centralized ellipsoid as
\begin{equation}
\label{eq:ellipsoid_def}
    \widetilde{E_k} = \left\{ x \in \mathbb{R}^n \mid \|A_k x\|_2 \leq \Delta_k \right\},
\end{equation}
where $\Delta_k=\frac{1}{|S_k|}\sum_{z_i \in S_k} ||z_i-c_k||_2$ is the average distance between the centroid $c_k$ and samples with label $k$ to enhance the numerical stability among known classes with different scales. This formulation eliminates the need for explicit orthogonality constraints while ensuring that \(\widetilde{E_k}\) remains an ellipsoid.

To demonstrate that $\widetilde{E_k}$ defined by Eq.\ref{eq:ellipsoid_def} is indeed an ellipsoid, we start by expanding the quadratic form induced by \(A_k\) to reveal the structure of the ellipsoid, which then leads naturally to the construction of a symmetric matrix and its spectral interpretation.
\paragraph{Quadratic Form Expansion:}  
Let \(A_k = [a_{m,i}] \in \mathbb{R}^{n \times n}\) be the learnable matrix and \(x = (x_1, \dots, x_n)^\top \in \mathbb{R}^n\). Then the squared norm \(\|A_k x\|_2^2\) can be expanded as:
\begin{equation}
 \label{eq:quadratic_expansion}
    \begin{aligned}
        \|A_k x\|_2^2 =& \sum_{i=1}^n \left( \sum_{k=1}^n a_{k,i} \right)^2 x_i^2 \\
        &+ 2\sum_{1 \leq i < j \leq n} \left( \sum_{k=1}^n a_{k,i}a_{k,j} \right) x_i x_j,
    \end{aligned}
\end{equation}
which reveals both the quadratic and cross terms.

\paragraph{Symmetric Matrix Construction:}  
We define a symmetric matrix \(\widehat{A_k} = [\widehat{a}_{i,j}]\) by
\begin{equation}
\label{eq:symmetric_matrix}
    \begin{aligned}
        \widehat{a}_{i,i} &= \sum_{m=1}^n a_{m,i}^2 \quad (\text{diagonal terms}), \\
        \widehat{a}_{i,j} = \widehat{a}_{j,i} &= \sum_{m=1}^n a_{m,i}a_{m,j} \quad (i \neq j),
    \end{aligned}
\end{equation}
which exactly captures the coefficients appearing in \eqref{eq:quadratic_expansion}. Therefore, we can rewrite
\begin{equation}
    \|A_k x\|_2^2 = x^\top \widehat{A_k} x.
\end{equation}

\paragraph{Spectral Interpretation:}  
The centralized ellipsoid can be equivalently expressed as
\begin{equation}
    \widetilde{E_k} = \{ x \in \mathbb{R}^n \mid x^\top \widehat{A_k} x \leq \Delta_k^2 \}.
\end{equation}
Since \(\widehat{A_k}\) is symmetric positive-definite, its spectral decomposition is given by $\widehat{A_k} = \sum_{i=1}^n \lambda_i\, d_i d_i^\top$, where \(\{d_i\}_{i=1}^n\) are orthonormal eigenvectors and \(\lambda_i > 0\) are the corresponding eigenvalues. This implies that the ellipsoid's axes are aligned with the eigenvectors \(d_1, \dots, d_n\) and have lengths $a_i = \frac{\Delta_k}{\sqrt{\lambda_i}}$.

It has been proved that $\widetilde{E_k}$ is a centralized ellipsoid whose axes are \(\{d_i\}_{i=1}^n\) and their lengths are \(\{a_i\}_{i=1}^n\) correspondingly, and thereby $E_k$ is also an ellipsoid parameterized as:
\begin{equation}
\label{eq:ellipsoid_def_final}
    E_k = \left\{ x \in \mathbb{R}^n \mid \|A_k (x-c_k)\|_2 \leq \Delta_k \right\}
\end{equation}
by \eqref{eq:centralized_ellipsoid} and \eqref{eq:ellipsoid_def}.

Finally, we define the affine transformation
\begin{equation}
    \varphi_k(z) = A_k (z - c_k)
\end{equation}
that map $E_k$ to a centralized ball. In particular,
\begin{equation}
    z \in E_k \iff \|\varphi_k(z)\|_2 \leq \Delta_k.
\end{equation}
This equivalence provides a convenient parametrization for points within the ellipsoid.

\subsubsection{Boundary Optimization Objectives}\label{sec:boundary_optimization}
To establish discriminative class boundaries, we design dual-directional boundary adjustment mechanisms through complementary loss functions. For known samples, boundaries are optimized to expand and encompass their distribution, while synthesized pseudo-negative samples induce boundary contraction in under-explored regions.

\textbf{Expansion Loss:} Given a labeled sample $z$ from class \( k \), we define the expansion loss as:
\begin{equation}
\label{eq:expansion_loss}
\mathcal{L}_{\mathrm{p},k}(z) := \max\left( \|\varphi_k(z)\|_2 - \Delta_k,\, 0 \right)
\end{equation}
This loss activates when \( z \) resides outside the current class boundary (\( \|\varphi_k(z)\|_2 > \Delta_k \)), creating gradient signals that \textit{expand} the boundary radially until it encloses the sample. Continuous application of this mechanism enables dynamic boundary adaptation to the inherent data distribution.

\textbf{Contraction Loss:} To mitigate over-generalization into regions without known samples, we generate pseudo-open samples as negative samples to encourage the boundaries to contract in those areas. The pseudo-open samples are synthesized by mixing \(P\) known samples with distinct labels~\citep{zhang2024multimodalclassificationoutofdistributiondetection}. Specifically, a pseudo-open sample is computed as
\begin{equation}
    \label{eq:pseudoOOD}
    z' = \sum_{i=1}^P \lambda_i z_i,\quad \text{with } \lambda \sim \mathrm{Dir}(\alpha),
\end{equation}
where \(P\) and \(\alpha\) are hyperparameters, and \(\{z_i\}_{i=1}^P\) are known samples with \(P\) different labels. The coefficients \(\{\lambda_i\}\) are drawn from a Dirichlet distribution (ensuring \(\sum_{i=1}^P \lambda_i = 1\) and \(\lambda_i \in [0,1]\)). Choosing \(\alpha < 1\) typically forces one coefficient to dominate, so that each pseudo-open sample tends to lie close to one known class while still exhibiting slight influences from the others.

The motivation behind the contraction loss is to push the boundary inward when a negative sample lies too close to or inside the boundary. For a negative sample \( \bm{z'}\) with respect to class \( k \), the negative loss is defined as: 
\begin{equation}
\label{eq:contraction_loss}
\mathcal{L}_{\mathrm{n},k}(z') = \begin{cases} 
        (\Delta_k - r_k(z')) + \beta, &  r_k(z') < \Delta_k \\
        \beta \mathrm{e}^{\Delta_k - r_k(z')}, & r_k(z') \geq \Delta_k
     \end{cases}
\end{equation}
where $r_k(z')=\|\varphi_k(z')\|_2$, \(\beta\) is a hyperparameter that controls the penalty strength. Specifically, when \(\|\varphi_k(z')\|_2 < \Delta_k\), the loss increases linearly as the negative sample penetrates deeper into the boundary, thus exerting a direct contraction force. When \(\|\varphi_k(z')\|_2 \geq \Delta_k\), the exponential term applies a softer penalty, ensuring a gradual enforcement of the boundary's inward contraction. The parameter \(\beta\) effectively balances these two regimes, with larger values of \(\beta\) inducing a stronger contraction effect on the boundary.

The final optimization objective $\mathcal{L}_{\mathrm{tot}}$ aggregates per-sample losses across all classes:
\begin{equation}
\label{eq:total_loss}
\mathcal{L}_{\mathrm{tot}} = \sum_{k=1}^K \left[\sum_{z\in\mathcal{B}\cap S_k}  \mathcal{L}_{\mathrm{p},k}(z)+\sum_{z' \in \mathcal{B}'} \mathcal{L}_{\mathrm{n},k}(z') \right]
\end{equation}
where \( \mathcal{B} \) denotes a training batch and \( \mathcal{B}' \) is a corresponding set of negative samples of the same size, generated by \(\mathcal{B}\). This formulation jointly coordinates boundary expansion and contraction through gradient signals from both positive and negative samples.

\subsection{Inference}
\label{sec:inference}

For a given sample \(z\), we first assign it to the nearest class based on the Euclidean distance to class centroids:
\begin{equation}
\label{eq:nearest_class}
j = \arg\min_{k \in \{1,\dots,K\}} \|z - c_k\|_2.
\end{equation}
Next, we check whether \(z\) falls within the ellipsoid boundary \(E_j\) by verifying
\begin{equation}
\label{eq:membership}
\|\varphi_j(z)\|_2 \le \Delta_j,
\end{equation}
which implies \(z \in E_j\). Otherwise, if
\begin{equation}
\label{eq:rejection}
\|\varphi_j(z)\|_2 > \Delta_j,
\end{equation}
$z$ is rejected as an open sample based on the high-confidence inference that \(z \notin \bigcup_{k=1}^{K} E_k\).

\begin{table*}[h]
\centering
{\footnotesize
\begin{tabular}{cccccccccccccc}
\toprule
   \multirow{2}{*}{Datasets}                       & \multirow{2}{*}{Methods}              &  & \multicolumn{2}{c}{KCR=25\%} &  & \multicolumn{2}{c}{KCR=50\%} &  & \multicolumn{2}{c}{KCR=75\%} &  & \multicolumn{2}{c}{Average} \\ 
                        &       &  & F1                & ACC              &  & F1              & ACC            &  & F1                   & ACC              &  & F1                & ACC \\ 
    \midrule

\multirow{10}{*}{Banking} & Softmax-MSP &  & 50.72 & 40.35 &  & 69.04 & 53.80 &  & 83.12 & 74.21 &  & 67.63 & 56.12 \\
     & DOC &  & 62.09 & 61.19 &  & 80.43 & 75.30 &  & \underline{86.78} & 81.23 &  & 76.43 & 72.57 \\
     & OpenMax &  & 56.19 & 53.00 &  & 74.40 & 65.01 &  & 84.88 & 77.87 &  & 71.82 & 65.29 \\
     & ADB &  & 73.21 & 81.50 &  & 81.09 & 79.35 &  & 85.61 & 80.72 &  & 79.97 & 80.52 \\
     & DA-ADB &  & 74.75 & 81.06 &  & 82.22 & 80.76 &  & 85.69 & 80.89 &  & \underline{80.89} & \underline{80.90} \\
     & KNNCL &  & \underline{77.30} & \underline{85.72} &  & 81.10 & \underline{82.61} &  & 82.22 & 77.35 &  & 80.21 & 81.89 \\
     & DE &  & 67.36 & 69.99 &  & 79.20 & 72.64 &  & 86.77 & 80.51 &  & 77.78 & 74.38 \\
     & CLAB &  & 72.85 & 79.04 &  & \underline{83.17} & 81.36 &  & 86.70 & \underline{81.92} &  & 80.91 & 80.77 \\
     & MOGB &  & 64.34 & 70.56 &  & 73.37 & 71.80 &  & 71.98 & 66.71 &  & 69.90 & 69.69 \\
     & \textit{EliDecide} &  & \textbf{77.75} & \textbf{85.81} &  & \textbf{83.74} & \textbf{82.90} &  & \textbf{86.80} & \textbf{81.97} &  & \textbf{82.76} & \textbf{83.56} \\
    \midrule

\multirow{10}{*}{OOS} & Softmax-MSP &  & 51.36 & 53.05 &  & 77.99 & 74.30 &  & 84.29 & 76.91 &  & 71.21 & 68.09 \\
     & DOC &  & 72.32 & 81.32 &  & 84.51 & 84.18 &  & \underline{89.56} & 86.90 &  & 82.13 & 84.13 \\
     & OpenMax &  & 64.89 & 72.33 &  & 80.03 & 80.99 &  & 74.58 & 77.06 &  & 73.17 & 76.79 \\
     & ADB &  & 79.18 & 89.03 &  & 85.77 & 87.21 &  & 88.83 & 86.85 &  & 84.59 & 87.70 \\
     & DA-ADB &  & 80.51 & 89.85 &  & \underline{86.62} & 88.84 &  & 88.53 & 87.47 &  & \underline{85.22} & 88.72 \\
     & KNNCL &  & \underline{82.05} & \textbf{92.95} &  & 84.51 & \underline{89.07} &  & 85.03 & 84.96 &  & 83.86 & \underline{88.99} \\
     & DE &  & 72.54 & 80.85 &  & 83.51 & 82.91 &  & 88.93 & 85.67 &  & 81.66 & 83.14 \\
     & CLAB &  & 78.24 & 87.25 &  & 86.40 & 88.44 &  & 89.42 & \underline{88.02} &  & 84.69 & 87.90 \\
     & MOGB &  & 72.34 & 82.15 &  & 75.31 & 77.86 &  & 71.52 & 71.47 &  & 73.06 & 77.16 \\
     & \textit{EliDecide} &  & \textbf{84.48} & \underline{92.57} &  & \textbf{88.57} & \textbf{90.56} &  & \textbf{90.21} & \textbf{88.98} &  & \textbf{87.75} & \textbf{90.70} \\
    \midrule

\multirow{10}{*}{StackOverflow} & Softmax-MSP &  & 38.13 & 27.69 &  & 62.30 & 49.50 &  & 76.17 & 69.99 &  & 58.87 & 49.06 \\
     & DOC &  & 45.84 & 38.88 &  & 64.88 & 53.81 &  & 78.14 & 72.01 &  & 62.95 & 54.90 \\
     & OpenMax &  & 45.37 & 38.49 &  & 69.68 & 63.44 &  & 79.82 & 74.62 &  & 64.96 & 58.85 \\
     & ADB &  & 80.19 & 87.19 &  & 84.72 & 85.60 &  & 86.10 & 82.98 &  & 83.67 & 85.26 \\
     & DA-ADB &  & \underline{83.44} & \underline{90.02} &  & \underline{86.77} & \underline{87.82} &  & 87.14 & 84.05 &  & \underline{85.78} & \underline{87.30} \\
     & KNNCL &  & 77.47 & 83.26 &  & 84.70 & 84.84 &  & 87.10 & 83.87 &  & 83.09 & 83.99 \\
     & DE &  & 60.86 & 60.12 &  & 75.92 & 71.58 &  & 84.18 & 79.78 &  & 73.65 & 70.49 \\
     & CLAB &  & 73.61 & 78.07 &  & 85.79 & 86.19 &  & \underline{87.64} & \underline{84.48} &  & 82.35 & 82.91 \\
     & MOGB &  & 70.91 & 76.78 &  & 79.13 & 77.91 &  & 82.47 & 78.06 &  & 77.50 & 77.58 \\
     & \textit{EliDecide} &  & \textbf{84.35} & \textbf{90.74} &  & \textbf{87.06} & \textbf{87.86} &  & \textbf{88.00} & \textbf{84.93} &  & \textbf{86.47} & \textbf{87.84} \\
    \bottomrule
\end{tabular}}
\caption{Results of the main experiments with different Known Class Ratio (KCR) settings on three datasets. To compare the generalization of methods, the average performances across three KCR settings are also reported in the last two columns for each dataset. The best performances are shown in \textbf{bold}, while the second-best are \underline{underlined}.}
\label{results}
\end{table*}

\section{Experiments}
\subsection{Experimental Settings}
\subsubsection{Datasets}
To validate the effectiveness of our method, we use three widely recognized datasets, including two intent datasets Banking~\cite{2020Efficient} and OOS~\cite{larson2019evaluation} and a question classification dataset StackOverflow~\cite{xu2015short} . Details are provided in the appendix. 

\subsubsection{Baselines}
We compare our method with the following open-world classification approaches: OpenMax \cite{bendale2016towards}, Softmax-MSP \cite{hendrycks2016baseline}, DOC \cite{shu2017doc}, ADB~\cite{zhang2021deep}, DA-ADB~\cite{zhang2023learning}, KNNCL~\cite{zhou2022knn}, DE~\cite{zhou2023two},  CLAB~\cite{Liu_Li_Mu_Yang_Xu_Wang_2023} and MOGB~\citep{Li_Ouyang_Pan_Zhang_Zhao_Xia_Yang_Wang_Li_2025}. Detailed descriptions of these baselines are provided in the appendix.

To ensure a fair comparison across different methods, we use BERT as the backbone model and conduct experiments on the open world classification platform TEXTOIR~\citep{zhang2021textoir} for all methods. We integrated the released implementations of DE, CLAB and MOGB into TEXTOIR. For other methods, the implementations included in TEXTOIR are used. 

\subsubsection{Evaluation Metrics}

Following the metric adopted in previous work \cite{zhang2021deep}, we use the average F1-score (F1) and the average accuracy (ACC) over all classes as metrics to evaluate the performance.

\subsubsection{Implementation Details}

Following previous work DA-ADB \cite{zhang2023learning}, we randomly select 25\%, 50\%, and 75\% of all classes as known classes with the remainder designated as unknown classes. The ratio of known classes selected is denoted as \textit{Known Class Ratio} (KCR). Samples from unknown classes are excluded from the training and validation sets but retained in the test set, ensuring no unknown samples are present in the training and validation processes. For each setting, we conduct five rounds of experiments using random seeds 0,1,2,3 and 4 respectively and report the averages. Further implementation details of each method are provided in the appendix. 

\subsection{Main Results}

Results are presented in Table~\ref{results}. Our method achieves the top results in almost all settings, demonstrating its effectiveness in enhancing both known and open intent classification. To assess generalization, we calculated the average performances across three KCR settings for each dataset. Our approach achieves 0.54\%-2.66\% improvement across all settings, highlighting the effectiveness and generalizability of the ellipsoid boundary and dual loss methodology in balancing empirical and open-space risks. Notably, compared to existing boundary-based methods ADB, DA-ADB and CLAB, our method exhibits performance gains specifically under 25\% and 50\% KCR settings. This improvement can be attributed to the ellipsoid boundary, which more effectively adapts to the presence of numerous unknown classes, resulting in a more precise decision boundary. This observation further suggests that our ellipsoid boundary design effectively improves the model's detection performance.

\section{Discussions}
To further validate the theory and demonstrate the effectiveness of our method, we conduct additional experiments and deep analysis from the following aspects: (1) the geometric advantage of ellipsoids comparing with balls, which is the primary novelty of our method; (2) ablation studies of major components; (3) the comparison with LLMs, showing the limitations of LLMs and supporting the value of our work. More discussions are stated in the appendix.

\subsection{Geometric Advantages of Ellipsoid}
\label{sec:compare_shapes}

To show the geometric advantages of ellipsoids, we compare them against balls with different radii. To avoid bias introduced by potentially suboptimal training procedures, we vary the radii by a hyperparameter called the \textit{Coverage Fraction} (CF) instead of determining them through training. For each known class $k$, the ball's radius is set as the distance from its centroid to the $\lfloor\mathrm{CF}|S_k|\rfloor$-th nearest training sample. Consequently, each ball boundary is designed to include a fraction CF of its class's training samples. 

\begin{figure}
    \centering
    \includegraphics[width=\linewidth]{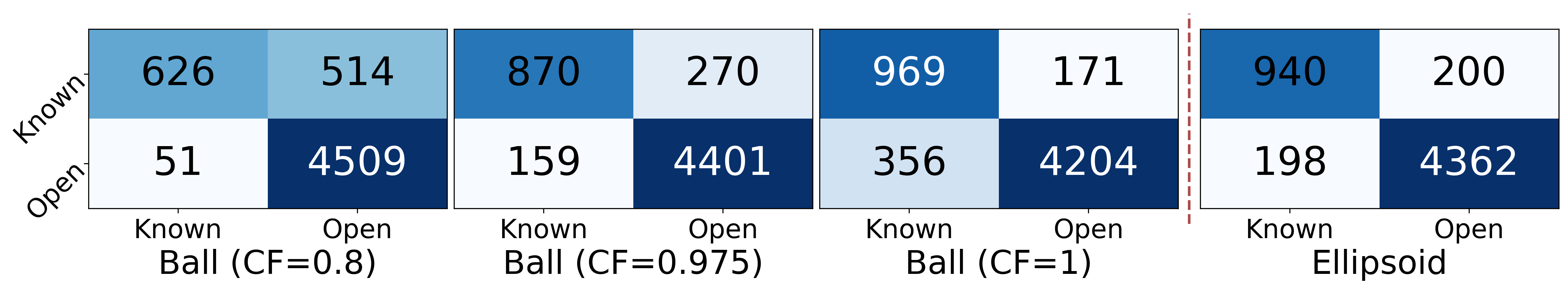}
    \caption{Coarse-grained confusion matrices with balls defined by different coverage fractions (CFs) and the ellipsoid (KCR=25\%, seed=0).}
    \label{fig:cm}
\end{figure}

Results are presented in the top section of Table~\ref{ablation_results}. Ball boundaries defined by CF fail to achieve optimal performance across different KCRs under the same hyperparameter, which highlights the necessity for adaptive training methods. Figure~\ref{fig:cm} illustrates the trade-off that a small CF causes excessive rejection of known samples as open. while a large CF fails to reject many open samples. Notably, even with KCR-specific CF tuning, balls underperform ellipsoids by 0.8\%-1.9\%. We attribute this gap to the inherent geometric constraint of balls, underscoring the advantage of the more flexible ellipsoidal shape. 

\subsection{Ablation Studies}

In this section we validate the effectiveness of our representation learning method and the negative loss function. The results on OOS are shown in Table~\ref{ablation_results}. 

\subsubsection{Representation Learning}
\label{finetune}

To explore the effect of our representation learning method, we discuss the effects of supervised contrastive learning (SCL) and normalization. Replacing SCL with a linear classifier and Softmax loss function reduces performance by 0.46\%-1.33\%, as the Softmax loss function solely focuses on distinguishing between different known classes but lacks intra-class compactness and inter-class separation. Removing feature normalization causes a 1.56\%-2.89\% drop at 25\% and 50\% known class ratios while flat at 75\%. These results suggest discriminative gains are stronger when known samples are scarce, likely because compact known-class distributions are crucial with limited information and abundant unknowns.

\subsubsection{Negative Loss Functions}
\label{pseudoOOD}

To validate the effectiveness of our proposed negative loss Eq.~\ref{eq:contraction_loss} which leverages generated pseudo-open samples, we compare it with the existing spherical boundary loss $\mathcal{L}_{\mathrm{ADB}}$~\citep{zhang2021deep} and two variants: CLAB loss~\citep{Liu_Li_Mu_Yang_Xu_Wang_2023} $\mathcal{L}_{\mathrm{CLAB}}$ and a variant ADB loss $\mathcal{L}_{\mathrm{ADB-Gen}}$ employing pseudo-open samples . $\mathcal{L}_{\mathrm{ADB}}$ employs known samples inside the boundary as negative samples and its variant $\mathcal{L}_{\mathrm{CLAB}}$ extends it by using extra known samples from other classes as negative samples. $\mathcal{L}_{\mathrm{ADB-Gen}}$ replaces known samples with generated pseudo open samples inside the boundary as negatives, which is equivalent to setting $\beta=0$ in Eq.~\ref{eq:contraction_loss} to nullify the exponential term. Formal definitions are provided in the appendix. 

Performance decreases sharply by 3\%-47\% particularlly under low KCR settings. We attribute this reduction to $\mathcal{L}_{\mathrm{ADB}}$'s reliance solely on intra-class information of known samples in a single class, which neglects critical inter-class relationships. While $\mathcal{L}_{\mathrm{CLAB}}$ mitigates this by using samples from other classes, improvements remain marginal. $\mathcal{L}_{\mathrm{ADB-Gen}}$ further incorporates the generated pseudo-open samples to model inter-class relationships but only activates those located inside the boundaries. In contrast, our method leverages the information across known classes to generate pseudo-open samples and activates all samples through the exponential term, better modeling the distribution of true open samples and leading to improved performance.

\begin{table}
    \centering
    \begin{tabular}{lccc}
        \toprule
    Methods    & 25\%                                      & 50\%                      &                 75\%                              \\  
\midrule
Ball (CF=0.8) & 72.29& 71.62 & 72.70\\
Ball (CF=0.9) & 79.25& 79.13& 80.24 \\
Ball (CF=0.95) & 82.24& 83.17& 84.89\\
Ball (CF=0.975) & \underline{83.20}& 85.05& 87.04 \\
Ball (CF=0.9875) & 82.49& 86.00& 88.25\\
Ball (CF=1) & 80.07& \underline{86.64}& \underline{89.28}\\
\midrule
Repre.: Softmax (w/o SCL) & 83.78        & 87.25           & 89.67         \\
Repre.: w/o Normalization    & 81.46        & 87.02        & \textbf{90.80}    \\ 
\midrule
Loss: $\mathcal{L}_{\mathrm{ADB}}$                & 37.42              & 59.21            & 74.19          \\
Loss: $\mathcal{L}_{\mathrm{CLAB}}$                & 42.49              & 60.67            & 74.53          \\
Loss: $\mathcal{L}_{\mathrm{ADB-Gen}}$ & 54.15 & 83.71& 86.52\\
\midrule
\textit{EliDecide}                           & \textbf{84.48} & \textbf{88.57}  & \underline{90.21} \\
\bottomrule
    \end{tabular} 
    \caption{Additional experimental results (F1) on OOS.}
    \label{ablation_results}
\end{table}

\subsection{Comparison with LLMs}

While LLMs have demonstrated strong capabilities across various natural language understanding tasks, the capabilities on open world classification has not been sufficiently investigated. In this section, we evaluate Llama~3~\citep{llama3modelcard} on open intent classification using multiple approaches and compare the results with our proposed method, highlighting the advantages of our approach using the relatively small-scaled pre-trained language model. 

Experiments are conducted on Banking~\citep{2020Efficient} with KCR=25\%. We evaluate combinations of different backbone models and detection methods. Regarding backbone models, we employ both the pre-trained model without modification and a finetuned variant adapted to training data for enhanced discriminability on known classes. For the detection methods, we implement two distinct approaches. First, we utilize the LLM as a direct generator through tailored prompts, requiring the model to either select one of $K$ known classes or classify the input as unknown (i.e. a $(K+1)$-way classification problem). This approach is evaluated under both zero-shot and five-shot settings. Second, we employ the LLM as a feature extractor analogous to BERT. The final token embeddings serve as sentence-level representations ($x_i$ in Eq.~\ref{eq:feature}), with subsequent predictions made by the ellipsoid boundaries. More implementation details are stated in the appendix.

\begin{table}
    \centering
    \begin{tabular}{lcccc}
        \toprule
        \multirow{2}{*}{Detection Methods} & \multicolumn{2}{c}{Base} & \multicolumn{2}{c}{Finetuned} \\
        & F1 & ACC & F1 & ACC \\
        \midrule
        End-to-End(zero-shot) & 38.03 & 23.60 & 45.13 & 26.00 \\
        End-to-End(five-shot) & 40.60 & 29.60 & 44.86 & 25.41 \\
        Boundary              & 24.84 & 54.24 & 73.76 & 83.70 \\
        \bottomrule
    \end{tabular}
    \caption{Results of the experiments on Llama 3 with different backbones and detection methods on Banking with KCR=25\%, where our proposed method's performance is F1=\textbf{77.75} and ACC=\textbf{85.81}. }
    \label{table:llm}
\end{table}

As shown in Table~\ref{table:llm}, finetuning on the known-class data improves classification performance. However, compared to specialized open intent detection methods, the end-to-end detection capability remains limited due to the absence of open samples during finetuning, consistent with the harmful \textit{finetuning-induced degradation of vocabulary} phenomenon revealed in~\citet{zhou-etal-2023-towards-open}. Notably, when substituting BERT with Llama 3 as the backbone, the results drop despite its larger scale. We attribute this to \textit{over-thinking} in deep transformer layers~\citep{zhou2023two}, which may impair open-world discriminability. These results show the unique advantages of our proposed method using a compact encoder and a designed detection method.

\section{Conclusion}

In this paper, we propose a novel ellipsoidal decision boundary framework EliDecide for open world classification. For each known class, we learn a nonsingular matrix that adaptively encodes both the ellipsoid's orientation and scale of its axes and optimize the boundaries via dual expansion and contraction losses. The geometric flexibility enables our method to jointly optimize empirical risk and open-space risk, achieving state-of-the-art performance across diverse benchmarks. Additional experiments validate each component, especially revealing the geometric advantage of ellipsoids over balls and the efficacy of our adaptive boundary methodology with a compact backbone encoder versus LLMs. Our results demonstrate that the ellipsoid's capacity to balance compactness and adaptability is critical for reconciling the dual challenges of preserving known-class fidelity while robustly rejecting unknown samples.

\section{Acknowledgments}
This paper is founded by National Natural Science Foundation of China (Grant No. 62173195) .

\bigskip

\appendix
\section{A. Detailed Experimental Settings}

This section documents the setup for all experiments, including hardware/software environments, dataset descriptions and primary hyperparameter configurations.

\subsection{Environments}
All experiments are executed on a server with an NVIDIA Tesla V100 (32GB VRAM) running Ubuntu 22.04.5 LTS. Key package versions include:
\begin{itemize}
    \item Python 3.11.4
    \item NumPy 1.25.1
    \item Scikit-learn 1.6.0
    \item PyTorch 2.0.1+cu117
    \item Tokenizers 0.19.1
    \item Transformers 4.42.4
\end{itemize}
\subsection{Dataset Details}

We evaluate our approach using three public benchmarks:

Banking \cite{2020Efficient} is set in a standalone banking scenario, featuring 77 intent labels and a total of 13,083 customer service queries. 

OOS \cite{larson2019evaluation} is an intent dataset with a rich variety of domains, containing 150 intents across 10 different domains. 

StackOverflow \cite{xu2015short} is a dataset in the field of computer science released by Kaggle.com, containing a total of 20,000 samples across 20 categories.

Further detailed statistics are presented in Table~\ref{dataset}.

\begin{table*}[h]
\centering
\begin{tabular}{llccccc}
\toprule
Dataset       & Classes & Training & Validation & Testing & Vocabulary & Length (max / mean) \\ \midrule
Banking       & 77      & 9,003       & 1,000        & 3,080     & 5,028      & 79 / 11.91          \\
OOS           & 150     & 15,000     & 3,000         & 5,700      & 8,288      & 28 / 8.31           \\
StackOverflow & 20      & 12,000     & 2,000        & 6,000     & 17,182     & 41 / 9.18           \\ \bottomrule
\end{tabular}
\caption{Detailed statistics of datasets. Columns Training, Validation and Testing are the numbers of sentences of each divided dataset.}
\label{dataset}
\end{table*}

\subsection{Implementation Details of Main Experiments}

We use the pre-trained uncased BERT model (12-layer transformer) implemented in PyTorch~\citep{wolf2020transformers} and conduct all experiments on TEXTOIR~\citep{zhang2021textoir} to ensure fair comparison. For all baseline methods, we adopt the hyperparameters from the released implementations. The hyperparameters for each method are summarized below.

\subsubsection{EliDecide (our proposed method)}

\begin{itemize}
    \item Feature dimension $n=768$;
    \item Representation learning rate $\mathrm{lr}=2\times 10^{-5}$;
    \item Boundary learning rate $\mathrm{lr}_b=0.001$-$0.002$;
    \item Temperature for contrastive learning $\tau=0.07$;
    \item Number of known samples mixed in pseudo-open sample generation $P=3$;
    \item Dirichlet distribution parameter in pseudo-open sample generation $\alpha=0.6$;
    \item Penalty strength $\beta=0.5$.
\end{itemize}
\subsubsection{Softmax-MSP~\citep{hendrycks2016baseline}}
This is a simple baseline using maximum softmax probabilities for open detection.

\begin{itemize}
    \item Feature dimension $n=768$;
    \item Learning rate $\mathrm{lr}=2\times 10^{-5}$;
    \item Threshold for open detection $\text{t}=0.5$.
\end{itemize}
\subsubsection{DOC~\citep{shu2017doc}}
DOC uses a 1-vs-rest sigmoid layer and Gaussian fitting for open detection.

\begin{itemize}
    \item Feature dimension $n=768$;
    \item Learning rate $\mathrm{lr}=2\times 10^{-5}$;
    \item Coefficient in threshold calculation $\alpha=3$.
\end{itemize}
\subsubsection{OpenMax~\citep{bendale2016towards}}

OpenMax extends softmax networks by replacing the penultimate layer with activation vectors for $(K+1)$-way classification probabilities.

\begin{itemize}
    \item Feature dimension $n=768$;
    \item Learning rate $\mathrm{lr}=2\times 10^{-5}$;
    \item Weibull tail size $s_t=20$;
    \item Number of activation classes $\alpha=10$;
    \item Threshold for open detection $\text{t}=0.5$.
\end{itemize}
\subsubsection{ADB~\citep{zhang2021deep}}

ADB learns representations via cross entropy and constructs spherical class boundaries. 

\begin{itemize} 
    \item Feature dimension $n=768$;
    \item Representation learning rate $\mathrm{lr}=2\times 10^{-5}$; 
    \item Boundary learning rate $\mathrm{lr}_b=0.05$.
\end{itemize}
\subsubsection{DA-ADB~\citep{zhang2023learning}}

This ADB variant introduces distance-aware representation learning to enhance discriminability. 

\begin{itemize} 
    \item Feature dimension $n=768$;
    \item Representation learning rate $\mathrm{lr}=2\times 10^{-5}$; 
    \item Boundary learning rate $\mathrm{lr}_b=0.05$; 
    \item Coefficient in cosine classifier $\alpha=4$ .
\end{itemize}
\subsubsection{KNNCL~\citep{zhou2022knn}}
KNNCL uses k-nearest neighbors as positive samples and a queue for negative samples to learn features and detects open samples with LOF. 

 \begin{itemize}
    \item Feature dimension $n=768$;
    \item Learning rate $\mathrm{lr}=1\times 10^{-5}$;
    \item Size of the queue $Q=7500$;
    \item Coefficient of contrastive loss $\lambda=0.1$;
    \item Temperature for contrastive learning $\tau=0.3$;
    \item Number of neighbors for contrastive learning $k=40$;
    \item Number of positive samples for contrastive learning $n_\text{pos}=3$;
    \item Number of neighbors for LOF $n_\text{neighbor}=25$;
    \item Contamination for LOF $c=0.05$.
\end{itemize}
\subsubsection{DE~\citep{zhou2023two}}
DE is a dynamic ensemble method with internal classifiers, enabling early inference exit during inference to accelerate processing while ensuring accuracy, and introduces a training strategy adapting to the dynamic inference behavior.

\begin{itemize}
    \item Feature dimension $n=768$;
    \item Learning rate $\mathrm{lr}=2\times 10^{-5}$;
    \item Coefficient of relevancy loss $\alpha=1$;
    \item Coefficient of redundancy loss $\beta=0.1$;
    \item Threshold of perturbation norm for adversarial samples $\epsilon=0.4$;
    \item Number of adversarial iterations $z=2$;
    \item Adversarial learning rate $\text{lr}_a=0.15$;
    \item Number of neighbors for LOF $n_\text{neighbor}=20$;
    \item Contamination for LOF $c=\mathrm{auto}$;
    \item Threshold for known samples $t_i=1.4$;
    \item Threshold for open samples $t_o=1.5$;
    \item Exponent of the layer number $\gamma=0.75$.
\end{itemize}
\subsubsection{CLAB~\citep{Liu_Li_Mu_Yang_Xu_Wang_2023}}

CLAB leverages K-Center Contrast Learning (KCCL) and a novel loss function with expanding and shrinking operations for boundary learning.

\begin{itemize}
    \item Feature dimension $n=768$;
    \item Representation learning rate $\mathrm{lr}=2\times 10^{-5}$; 
    \item Boundary learning rate $\mathrm{lr}_b=0.05$; 
    \item Number of positive samples $K=5$;
    \item Number of negative samples $M=1$;
    \item Coefficient of KCCL loss $\lambda=0.25$;
    \item Activation threshold of expending $e=0.5$;
    \item Activation threshold of shrinking $s=0.1$;
    \item Coefficient of expending and shrinking loss $\eta=0.1$.
\end{itemize}

\subsubsection{MOGB~\citep{Li_Ouyang_Pan_Zhang_Zhao_Xia_Yang_Wang_Li_2025}}

MOGB constructs multiple spherical boundaries per class and optimizes the representation via a two-stage iterative procedure. 

\begin{itemize}
    \item Feature dimension $n=768$;
    \item Learning rate for cross entropy loss $\mathrm{lr}=2\times 10^{-5}$;
    \item Learning rate for granular ball loss $\mathrm{lr_gb}=1\times 10^{-4}$;
    \item Purity limit for splitting during training $p_l=0.9$;
    \item Purity limit for selection during training $p_t=1$;
    \item Sample count limit for splitting during training $n_l=5$ (for Banking \& OOS) and $n_l=15$ (for StackOverflow);
    \item Sample count limit for selection during training $n_t=1$;
    \item Purity limit for splitting during evaluation $p_l'=1$;
    \item Purity limit for selection during evaluation $p_t'=0.9$;
    \item Sample count limit for splitting during evaluation $n_l'=5$ (for Banking \& OOS) and $n_l'=15$ (for StackOverflow);
    \item Sample count limit for selection during evaluation $n_t'=5$ (for Banking \& OOS) and $n_t'=15$ (for StackOverflow);
    \item Ball radius: maximum distance from centroid to any sample assigned to the ball.
\end{itemize}

\subsection{Implementation Details of Additional Experiments}

\subsubsection{Negative Loss Functions}
The three negative loss functions compared in this work are defined as the following.
\begin{equation}
    \label{eq:ADB_negative_loss}
    \mathcal{L}_{\mathrm{n},k}^{\mathrm{ADB}}(z) := \max\left( \Delta_k - r_k(z), 0 \right).
\end{equation}
This term activates when sample $z$ lies inside the decision boundary, compressing the boundary.
\begin{equation}
    \label{eq:CLAB_negative_loss}
    \begin{aligned}
    \mathcal{L}_{\mathrm{n},k}^{\mathrm{CLAB}}(z) &:= \mathcal{L}_{\mathrm{n},k}^{\mathrm{ADB}}(z)\\
    &+ \eta\max\left( (\Delta_k+s) - r_k(z^-), 0 \right)\\
        &+\eta\max\left(r_k(z^-)-(\Delta_k+e),\, 0 \right),
    \end{aligned}
\end{equation}
where $z^-$ denotes a randomly selected negative sample from the batch, and $\eta,e,s$ are hyperparameters matching those in the CLAB baseline. 
\begin{equation}
    \label{eq:ADBGen_negative_loss}
    \mathcal{L}_{\mathrm{n},k}^{\mathrm{ADB-Gen}}(z') := \max\left( \Delta_k - r_k(z'), 0 \right),
\end{equation}
where $z'$ represents a pseudo-open sample generated using our proposed method's procedure.

\subsubsection{Comparison with LLMs}

We use Meta-Llama-3-8B-Instruct~\citep{llama3modelcard} for these experiments. For finetuning, we employ  Low-Rank Adaptation~\citep{hu2021loralowrankadaptationlarge} via LlamaFactory~\citep{zheng2024llamafactoryunifiedefficientfinetuning}, using the training and validation data prepared identically to the main experiments. The model is finetuned with autoregressive loss under a tailored prompt to generate correct labels. We set the learning rate to $5\times 10^{-4}$ athe best model based on minimal validation loss for inference. Below is an example prompt (used for both finetuning and inference), which includes the task description, user input and known class labels:

\begin{tcolorbox}[title=\text{Prompt Example} , colback=white, colframe=black, fonttitle=\bfseries]
\fontsize{9}{10.8}\selectfont
You are an intent classification assistant in the banking domain. Based on the content provided, identify the main purpose or objective behind the user's request. You will be provided with a sentence containing the user's request and a list of known intents. Please note that it is possible the user's intent does not match any of the known intents in the given list,in which case the answer should be the special label \texttt{'\textless{}UNK\textgreater{}'}. The list of known intents is: \texttt{[pending\_top\_up, why\_verify\_identity, top\_up\_failed, supported\_cards\_and\_currencies, exchange\_via\_app, declined\_card\_payment, unable\_to\_verify\_identity, exchange\_rate, passcode\_forgotten, verify\_my\_identity} \textit{(Other 9 labels ignored)}\texttt{]}, The user's request is:
\texttt{"Could you help my figure out the exchange fee?"}. Your output should be exactly one label: either one of the labels above, or \texttt{'\textless{}UNK\textgreater{}'} if none apply. Do not include any extra characters.
\end{tcolorbox}

For end-to-end detection, we use the original prompt above for the zero-shot setting. In the five-shot setting, we prepend five randomly sampled training examples to the prompt.

For ellipsoid boundary detection, we extract features using the same prompt and train \textit{EliDecide} with hyperparameters consistent with the main experiment.

\begin{table*}[h]
\centering
{\footnotesize
\begin{tabular}{ccccccccccc}
\toprule
   \multirow{2}{*}{Datasets}                       & \multirow{2}{*}{Methods}              &  & \multicolumn{2}{c}{KCR=25\%} &  & \multicolumn{2}{c}{KCR=50\%} &  & \multicolumn{2}{c}{KCR=75\%}  \\ 
                        &       &  & F1                & ACC              &  & F1              & ACC            &  & F1                   & ACC               \\ 
    \midrule

    \multirow{10}{*}{Banking} & Softmax-MSP &  & 50.72\ \tiny{4.42} & 40.35\ \tiny{52.69} &  & 69.04\ \tiny{1.96} & 53.80\ \tiny{4.45} &  & 83.12\ \tiny{1.36} & 74.21\ \tiny{4.13} \\
     & DOC &  & 62.09\ \tiny{33.70} & 61.19\ \tiny{65.83} &  & 80.43\ \tiny{0.78} & 75.30\ \tiny{3.10} &  & \underline{86.78}\ \tiny{0.35} & 81.23\ \tiny{1.37} \\
     & OpenMax &  & 56.19\ \tiny{2.11} & 53.00\ \tiny{15.12} &  & 74.40\ \tiny{1.40} & 65.01\ \tiny{1.58} &  & 84.88\ \tiny{0.51} & 77.87\ \tiny{1.04} \\
     & ADB &  & 73.21\ \tiny{10.03} & 81.50\ \tiny{7.30} &  & 81.09\ \tiny{1.91} & 79.35\ \tiny{1.18} &  & 85.61\ \tiny{0.32} & 80.72\ \tiny{0.54} \\
     & DA-ADB &  & 74.75\ \tiny{10.04} & 81.06\ \tiny{6.75} &  & 82.22\ \tiny{0.68} & 80.76\ \tiny{0.87} &  & 85.69\ \tiny{0.46} & 80.89\ \tiny{2.36} \\
     & KNNCL &  & \underline{77.30}\ \tiny{11.73} & \underline{85.72}\ \tiny{15.83} &  & 81.10\ \tiny{2.36} & \underline{82.61}\ \tiny{2.14} &  & 82.22\ \tiny{1.73} & 77.35\ \tiny{2.51} \\
     & DE &  & 67.36\ \tiny{11.94} & 69.99\ \tiny{23.29} &  & 79.20\ \tiny{2.90} & 72.64\ \tiny{3.86} &  & 86.77\ \tiny{0.34} & 80.51\ \tiny{1.74} \\
     & CLAB &  & 72.85\ \tiny{8.66} & 79.04\ \tiny{10.77} &  & \underline{83.17}\ \tiny{0.96} & 81.36\ \tiny{1.30} &  & 86.70\ \tiny{1.57} & \underline{81.92}\ \tiny{3.78} \\
          & MOGB &  & 72.34\ \tiny{2.08} & 82.15\ \tiny{11.67} &  & 75.31\ \tiny{4.21} & 77.86\ \tiny{7.81} &  & 71.52\ \tiny{8.28} & 71.47\ \tiny{3.21} \\
     & \textit{EliDecide} &  & \textbf{77.75}\ \tiny{6.14} & \textbf{85.81}\ \tiny{3.27} &  & \textbf{83.74}\ \tiny{1.71} & \textbf{82.90}\ \tiny{0.96} &  & \textbf{86.80}\ \tiny{0.76} & \textbf{81.97}\ \tiny{2.60} \\
    \midrule 
    \multirow{10}{*}{OOS} & Softmax-MSP &  & 51.36\ \tiny{12.36} & 53.05\ \tiny{39.86} &  & 77.99\ \tiny{4.56} & 74.30\ \tiny{18.48} &  & 84.29\ \tiny{10.57} & 76.91\ \tiny{34.42} \\
     & DOC &  & 72.32\ \tiny{7.00} & 81.32\ \tiny{6.93} &  & 84.51\ \tiny{0.03} & 84.18\ \tiny{0.45} &  & \underline{89.56}\ \tiny{0.29} & 86.90\ \tiny{0.63} \\
     & OpenMax &  & 64.89\ \tiny{6.61} & 72.33\ \tiny{13.54} &  & 80.03\ \tiny{0.46} & 80.99\ \tiny{1.64} &  & 74.58\ \tiny{41.34} & 77.06\ \tiny{15.36} \\
     & ADB &  & 79.18\ \tiny{5.55} & 89.03\ \tiny{3.86} &  & 85.77\ \tiny{0.03} & 87.21\ \tiny{0.28} &  & 88.83\ \tiny{0.12} & 86.85\ \tiny{0.30} \\
     & DA-ADB &  & 80.51\ \tiny{8.47} & 89.85\ \tiny{6.30} &  & \underline{86.62}\ \tiny{0.81} & 88.84\ \tiny{0.48} &  & 88.53\ \tiny{0.20} & 87.47\ \tiny{0.45} \\
     & KNNCL &  & \underline{82.05}\ \tiny{2.19} & \textbf{92.95}\ \tiny{0.93} &  & 84.51\ \tiny{1.63} & \underline{89.07}\ \tiny{0.50} &  & 85.03\ \tiny{0.40} & 84.96\ \tiny{0.22} \\
     & DE &  & 72.54\ \tiny{11.35} & 80.85\ \tiny{11.52} &  & 83.51\ \tiny{0.27} & 82.91\ \tiny{0.70} &  & 88.93\ \tiny{0.18} & 85.67\ \tiny{0.53} \\
     & CLAB &  & 78.24\ \tiny{5.71} & 87.25\ \tiny{8.38} &  & 86.40\ \tiny{0.18} & 88.44\ \tiny{0.75} &  & 89.42\ \tiny{0.15} & \underline{88.02}\ \tiny{0.06} \\
          & MOGB &  & 72.34\ \tiny{2.08} & 82.15\ \tiny{11.67} &  & 75.31\ \tiny{4.21} & 77.86\ \tiny{7.81} &  & 71.52\ \tiny{8.28} & 71.47\ \tiny{3.21} \\
     & \textit{EliDecide} &  & \textbf{84.48}\ \tiny{4.81} & \underline{92.57}\ \tiny{3.44} &  & \textbf{88.57}\ \tiny{0.64} & \textbf{90.56}\ \tiny{0.54} &  & \textbf{90.21}\ \tiny{0.08} & \textbf{88.98}\ \tiny{0.38} \\
    \midrule 
    \multirow{10}{*}{StackOverflow} & Softmax-MSP &  & 38.13\ \tiny{21.11} & 27.69\ \tiny{8.53} &  & 62.30\ \tiny{1.73} & 49.50\ \tiny{0.60} &  & 76.17\ \tiny{0.65} & 69.99\ \tiny{0.22} \\
     & DOC &  & 45.84\ \tiny{12.66} & 38.88\ \tiny{29.74} &  & 64.88\ \tiny{3.64} & 53.81\ \tiny{8.34} &  & 78.14\ \tiny{1.01} & 72.01\ \tiny{1.16} \\
     & OpenMax &  & 45.37\ \tiny{18.61} & 38.49\ \tiny{49.74} &  & 69.68\ \tiny{5.77} & 63.44\ \tiny{15.56} &  & 79.82\ \tiny{0.70} & 74.62\ \tiny{1.76} \\
     & ADB &  & 80.19\ \tiny{7.71} & 87.19\ \tiny{6.94} &  & 84.72\ \tiny{2.32} & 85.60\ \tiny{2.57} &  & 86.10\ \tiny{0.66} & 82.98\ \tiny{0.72} \\
     & DA-ADB &  & \underline{83.44}\ \tiny{7.82} & \underline{90.02}\ \tiny{4.92} &  & \underline{86.77}\ \tiny{1.79} & \underline{87.82}\ \tiny{1.74} &  & 87.14\ \tiny{0.54} & 84.05\ \tiny{0.84} \\
     & KNNCL &  & 77.47\ \tiny{19.17} & 83.26\ \tiny{26.29} &  & 84.70\ \tiny{8.77} & 84.84\ \tiny{13.68} &  & 87.10\ \tiny{0.53} & 83.87\ \tiny{0.61} \\
     & DE &  & 60.86\ \tiny{13.68} & 60.12\ \tiny{72.48} &  & 75.92\ \tiny{7.15} & 71.58\ \tiny{14.82} &  & 84.18\ \tiny{2.79} & 79.78\ \tiny{4.42} \\
     & CLAB &  & 73.61\ \tiny{7.99} & 78.07\ \tiny{18.22} &  & 85.79\ \tiny{2.05} & 86.19\ \tiny{3.14} &  & \underline{87.64}\ \tiny{0.43} & \underline{84.48}\ \tiny{0.88} \\
          & MOGB &  & 70.91\ \tiny{31.10} & 76.78\ \tiny{40.55} &  & 79.13\ \tiny{8.56} & 77.91\ \tiny{9.13} &  & 82.47\ \tiny{5.22} & 78.06\ \tiny{6.69} \\
     & \textit{EliDecide} &  & \textbf{84.35}\ \tiny{5.68} & \textbf{90.74}\ \tiny{3.98} &  & \textbf{87.06}\ \tiny{2.81} & \textbf{87.86}\ \tiny{3.33} &  & \textbf{88.00}\ \tiny{1.29} & \textbf{84.93}\ \tiny{2.02} \\
    \bottomrule
\end{tabular}}
\caption{Results of the main experiments with different Known Class Ratio (KCR) settings on three datasets.  Average performance is shown in regular font, with variances in small font. The best average performances are shown in \textbf{bold}, while the second-best are \underline{underlined}.}
\label{table:variance}
\end{table*}
\begin{table}
    \centering
    \begin{tabular}{lcccc}
        \toprule
        \multirow{2}{*}{Methods} & \multicolumn{2}{c}{F1} & \multicolumn{2}{c}{ACC} \\
        & t-Stat. & p-Value & t-Stat. & p-Value \\
        \midrule
    Softmax-MSP & 4.22 & 2.92e-03 & 4.82 & 1.33e-03 \\
    DOC & 2.85 & 2.13e-02 & 2.98 & 1.77e-02 \\
    OpenMax & 4.36 & 2.42e-03 & 4.14 & 3.26e-03 \\
    ADB & 6.07 & 3.00e-04 & 8.59 & 2.60e-05 \\
    DA-ADB & 4.43 & 2.21e-03 & 3.75 & 5.66e-03 \\
    KNNCL & 4.73 & 1.48e-03 & 2.79 & 2.37e-02 \\
    DE & 3.31 & 1.08e-02 & 3.84 & 4.97e-03 \\
    CLAB & 2.51 & 3.63e-02 & 2.56 & 3.36e-02 \\
    MOGB & 7.31 & 8.34e-05 & 6.88 & 1.28e-04 \\
        \bottomrule
    \end{tabular}
    \caption{Results of paired t-tests comparing our proposed method with baseline methods on F1 and ACC. For each baseline, the t-statistic and p-value are reported for both metrics.}
    \label{table:ttest}
\end{table}
\begin{table}
    \centering
    \begin{tabular}{lccc}
        \toprule
    Methods    & 25\%                                      & 50\%                      &                 75\%                              \\  
\midrule
LOF (con=0.05)                 & \underline{83.80}     & \underline{84.93}              & \underline{86.21}       \\
LOF (con=0.1)                         & 80.90                 & 81.17               & 82.11       \\
LOF (con=0.2)                         & 74.71          & 73.95         & 74.94    \\
LOF (con=0.3)                         & 68.03         & 67.02         & 67.39  \\
\midrule
\textit{EliDecide}                           & \textbf{84.48} & \textbf{88.57}  & \textbf{90.21} \\
\bottomrule
    \end{tabular} 
    \caption{Experimental Results (F1) of LOF on OOS.}
    \label{table:LOF}
\end{table}

\section{B. Supplementary Results and Discussions}

This section presents detailed statistical analyses of the main experimental results and provides additional experiments to offer deeper insights into our method. All experiments (including those in the main paper and this appendix) are executed for five times with random seeds 0, 1, 2, 3 and 4 and the average performances are reported. 

\subsection{Statistical Analysis of Main Experimental Results}

Table~\ref{table:variance} presents the variances across the five runs with different seeds. Our method achieves the top results in almost all settings while maintaining acceptable sensitivity to random seeds. We perform pairwise t-tests between our proposed method and each baseline respectively and report the t-statistics and p-values in Table~\ref{table:ttest}. For all baselines, the proposed method demonstrates statistically significant improvements ($p < 0.05$) in both metrics, with consistently positive t-statistics indicating superior performance.

\subsection{Comparison with LOF}
\label{LOF}

To demonstrate the advantage of the adaptive boundary approach, we compare our method against the conventional outlier detection algorithm, LOF~\cite{Breunig_Kriegel_Ng_Sander_2000}, under various settings for \textit{contamination}. The results are shown in Table~\ref{table:LOF}. The efficacy of LOF is lower than out method by 0.55\%-4.00\% and dependent on the hyperparameter choice, thereby highlighting the comparative advantage of our adaptive boundary method, which is hyperparameter-free.

\subsection{More Results of LLM}
\begin{table}
    \centering
    \begin{tabular}{lcccc}
        \toprule
        \multirow{2}{*}{Detection Methods} & \multicolumn{2}{c}{Base} & \multicolumn{2}{c}{Finetuned} \\
        & F1 & ACC & F1 & ACC \\
        \midrule
        End-to-End(zero-shot) & 44.37 & 42.16 & 78.29 & 70.03 \\
        End-to-End(five-shot) & 47.34 & 44.60 & 78.23 & 69.77 \\
        Boundary              & 24.19 & 25.27 & 79.26 & 73.73 \\
        \bottomrule
    \end{tabular}
    \caption{Results of the experiments on Llama 3 with different backbones and detection methods on Banking with KCR=75\%, where our proposed method's performance is F1=\textbf{86.80} and ACC=\textbf{81.97}. }
    \label{table:appendix_llm}
\end{table}

To further validate our conclusion about the LLM's performance, we conduct the experiments on Banking with KCR=75\% and present the results in Table~\ref{table:appendix_llm}. While the end-to-end method shows improved performance compared to the KCR=25\% setting (reported in the main paper), it still underperforms our boundary-based method. The best LLM results remain lower than our proposed BERT-based method, which aligns with and reinforces the conclusion presented in the main body of the paper.

\subsection{Hyperparameters Stability}

\begin{figure*}
    \centering
    \setlength{\floatsep}{5pt} 
    \setlength{\intextsep}{5pt}

    \begin{subfigure}[b]{0.33\textwidth}
        \includegraphics[width=\textwidth]{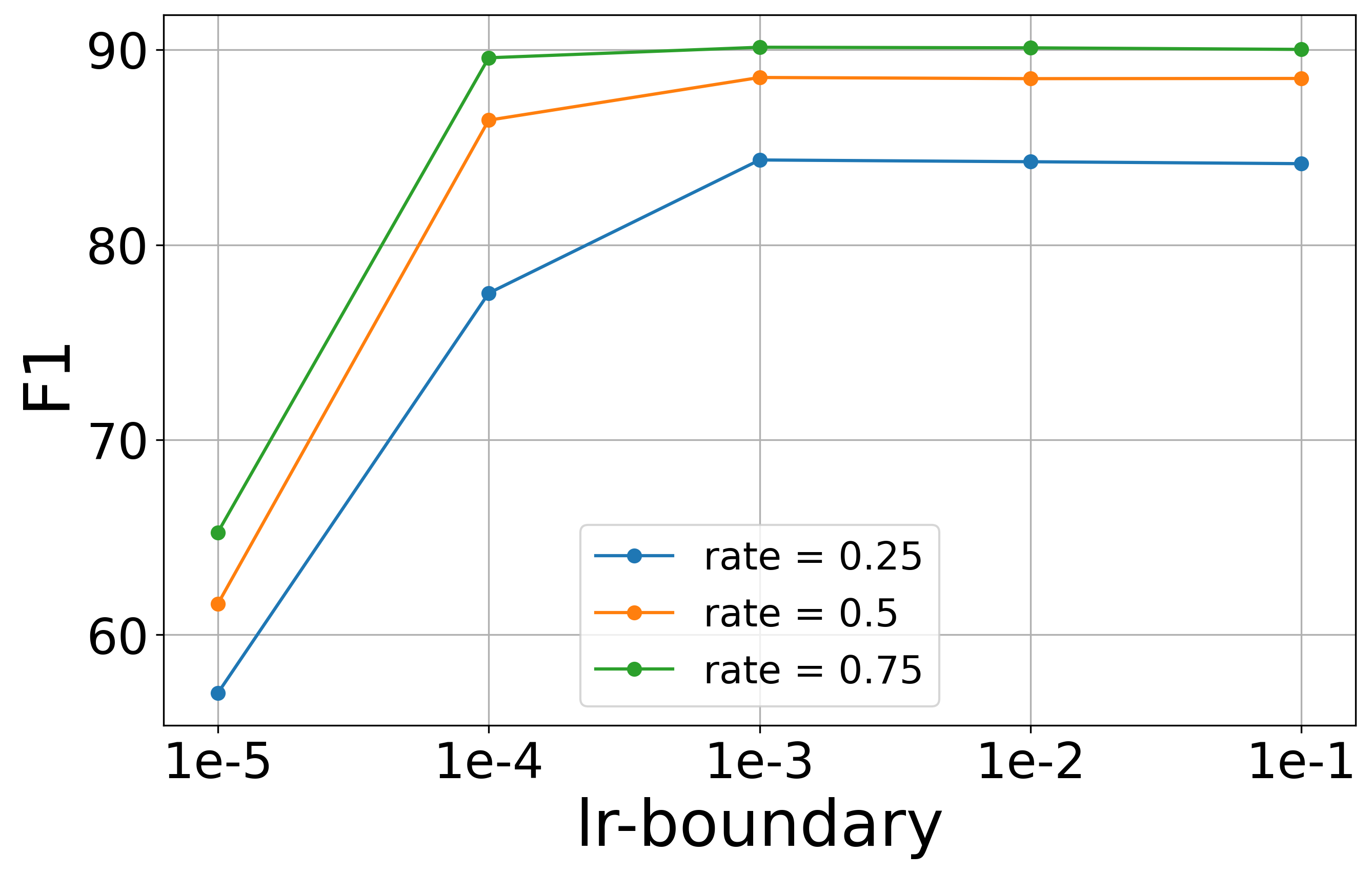}
        \caption{F1 for Learning Rate}
    \end{subfigure}
    \begin{subfigure}[b]{0.33\textwidth}
        \includegraphics[width=\textwidth]{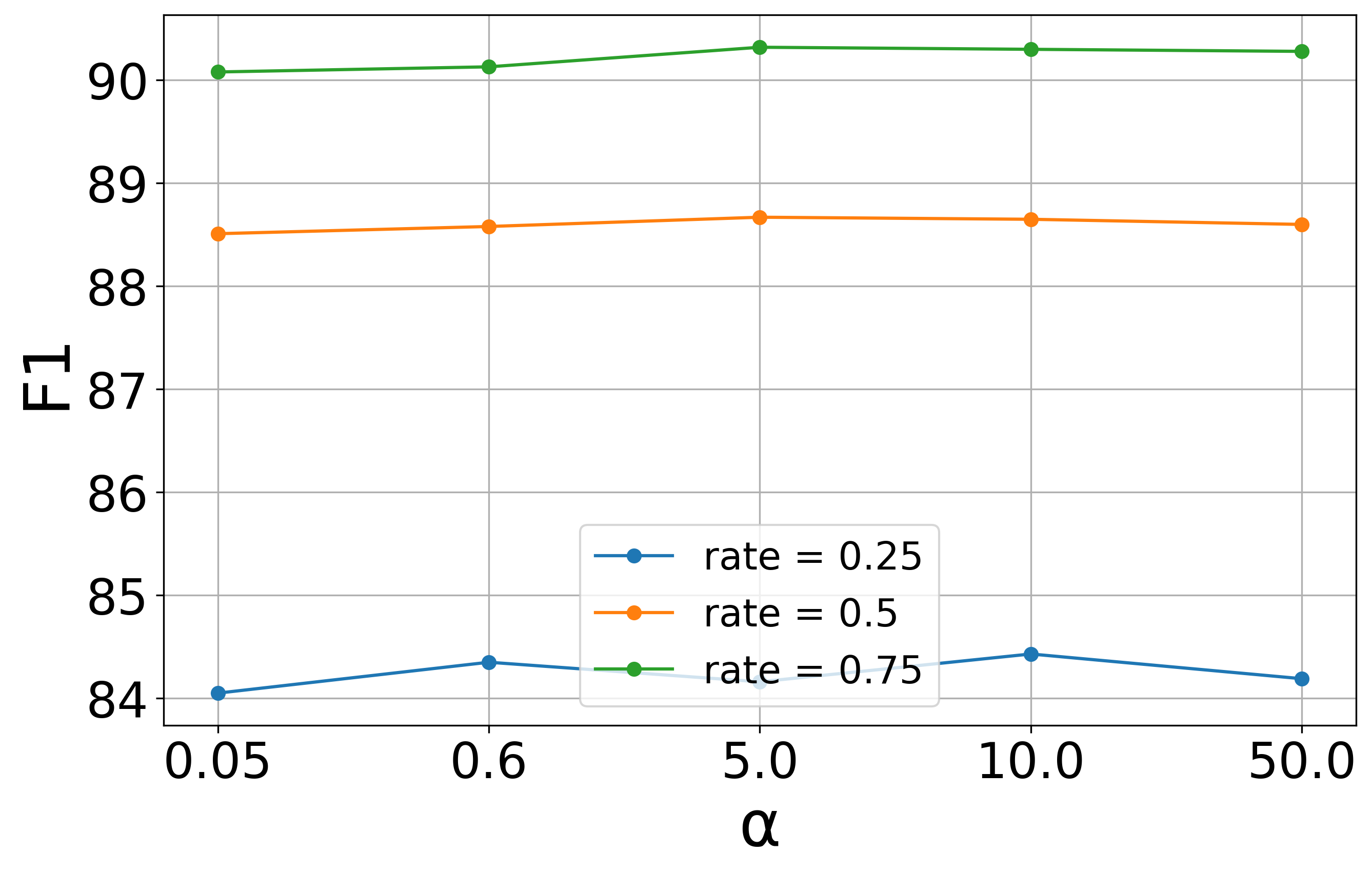}
        \caption{F1 for \(\alpha\)}
    \end{subfigure}
    \begin{subfigure}[b]{0.33\textwidth}
        \includegraphics[width=\textwidth]{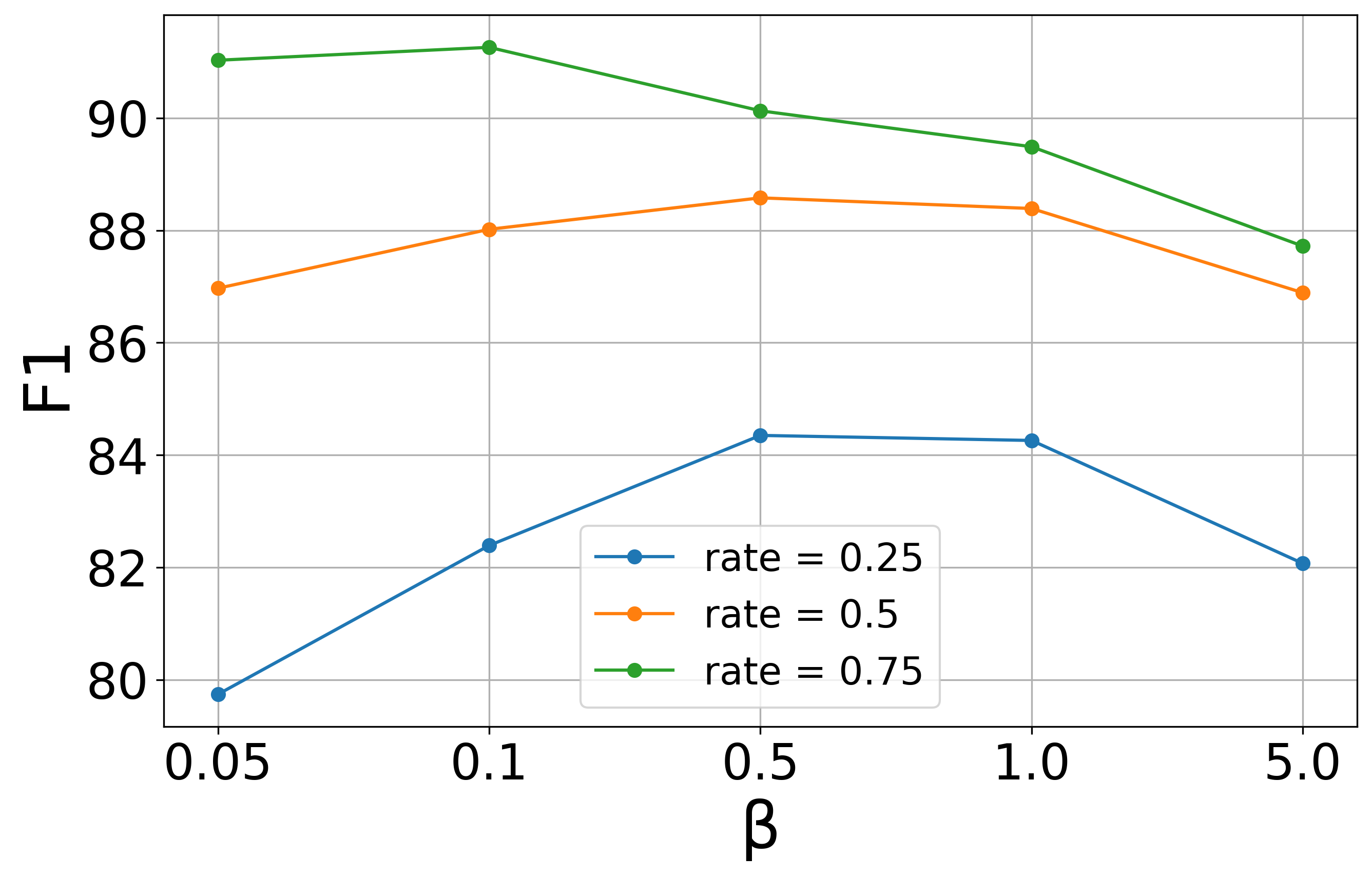}
        \caption{F1 for \(\beta\)}
    \end{subfigure}
    
    \begin{subfigure}[b]{0.33\textwidth}
        \includegraphics[width=\textwidth]{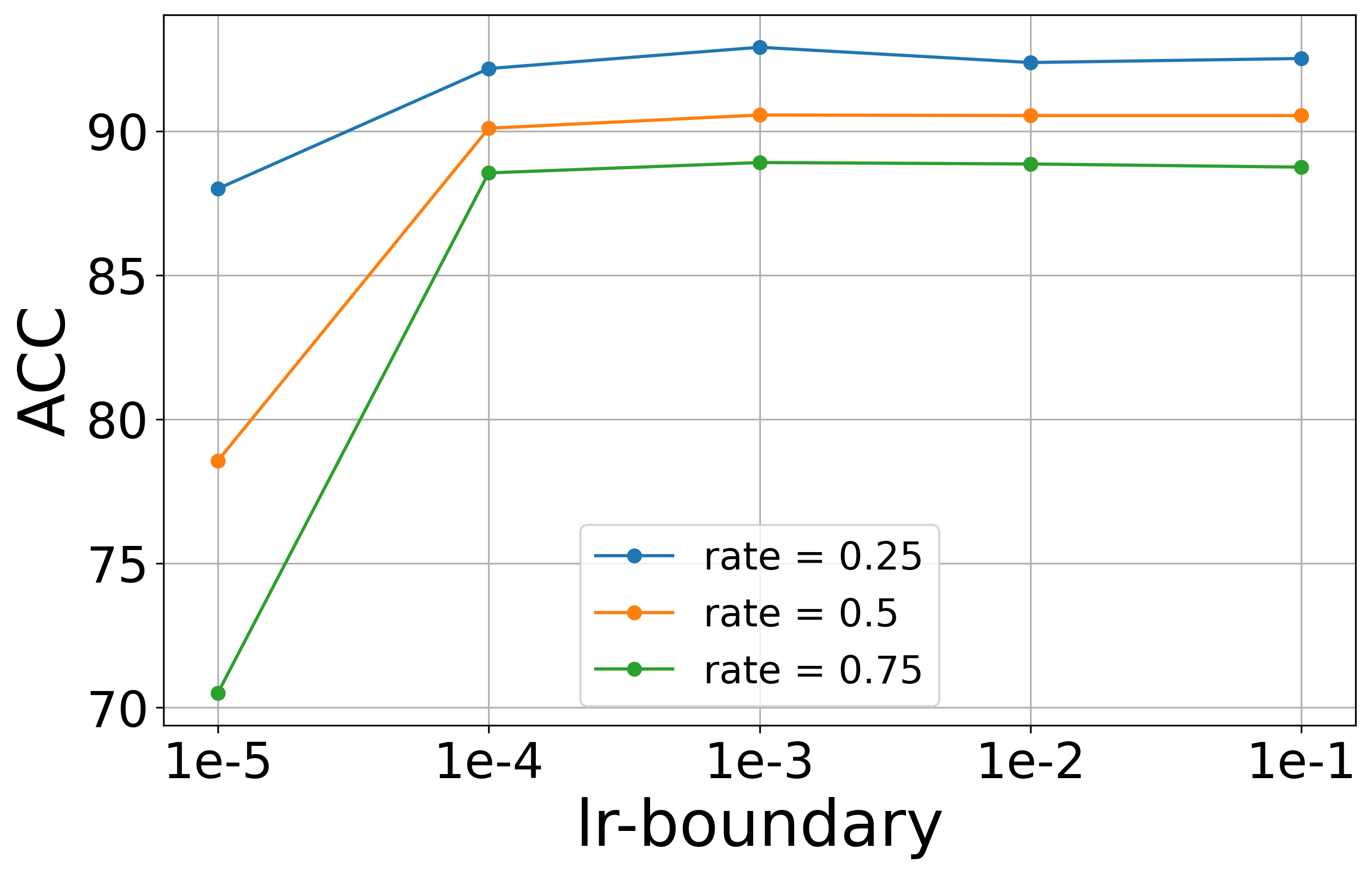}
        \caption{Acc for Learning Rate}
    \end{subfigure}
    \begin{subfigure}[b]{0.33\textwidth}
        \includegraphics[width=\textwidth]{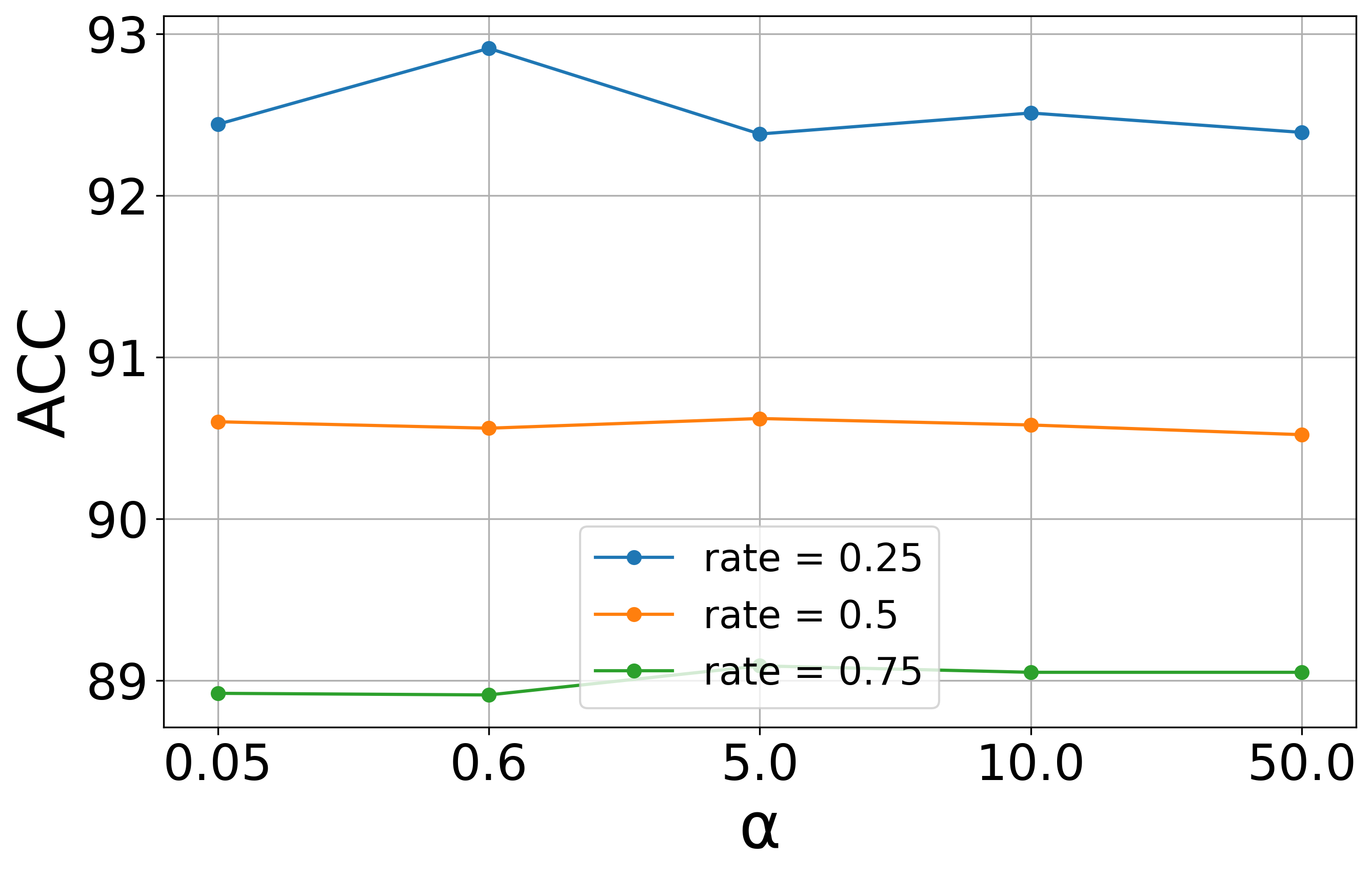}
        \caption{Acc for \(\alpha\)}
    \end{subfigure}
    \begin{subfigure}[b]{0.33\textwidth}
        \includegraphics[width=\textwidth]{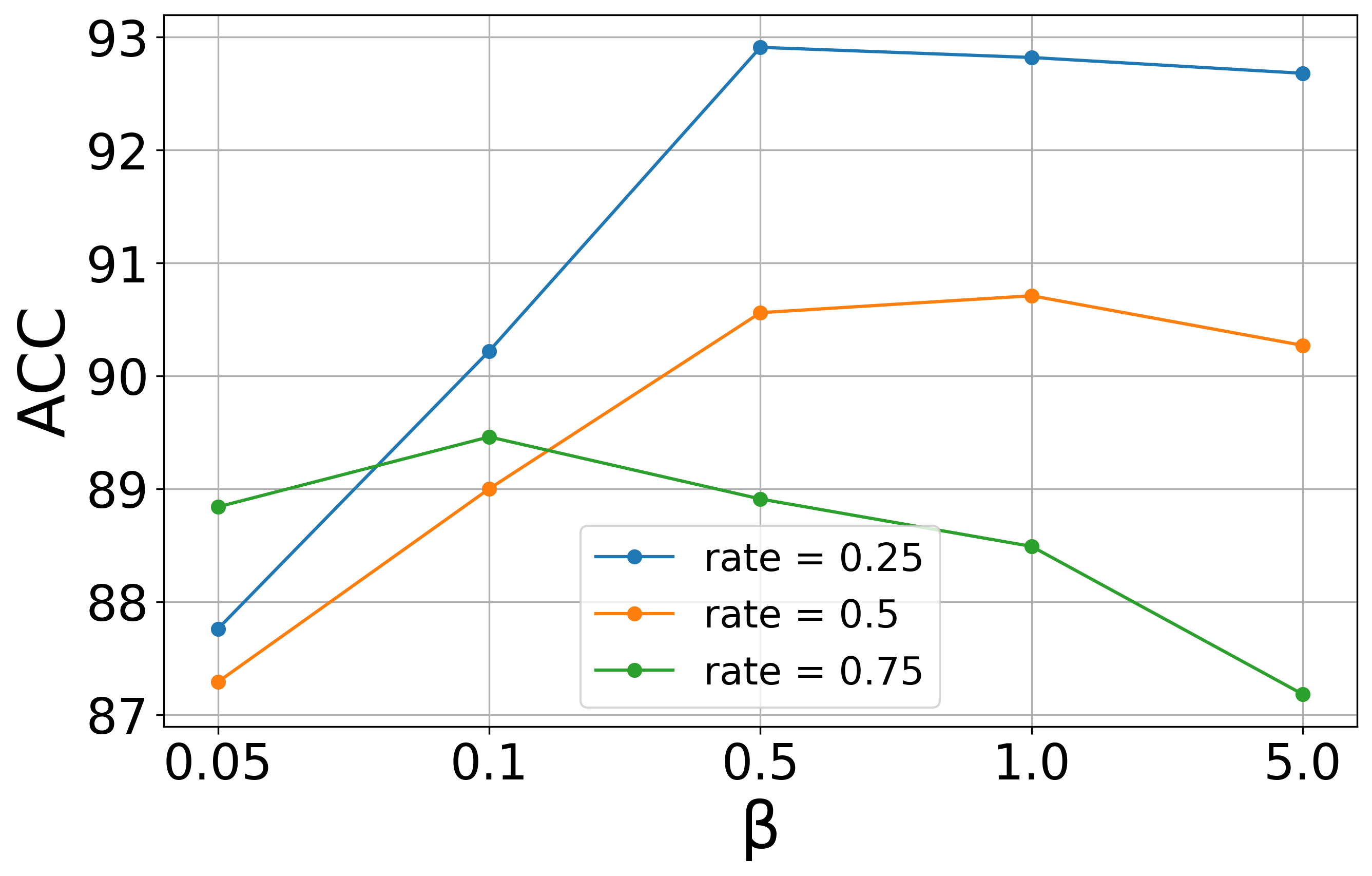}
        \caption{Acc for \(\beta\)}
    \end{subfigure}
    
    \caption{Comparison of F1 and Accuracy for different hyperparameters on OOS.}
    \label{fig:all_images_oos}
\end{figure*}

\begin{figure*}
    \centering
    \setlength{\floatsep}{5pt} 
    \setlength{\intextsep}{5pt} 

    \begin{subfigure}[b]{0.33\textwidth}
        \includegraphics[width=\textwidth]{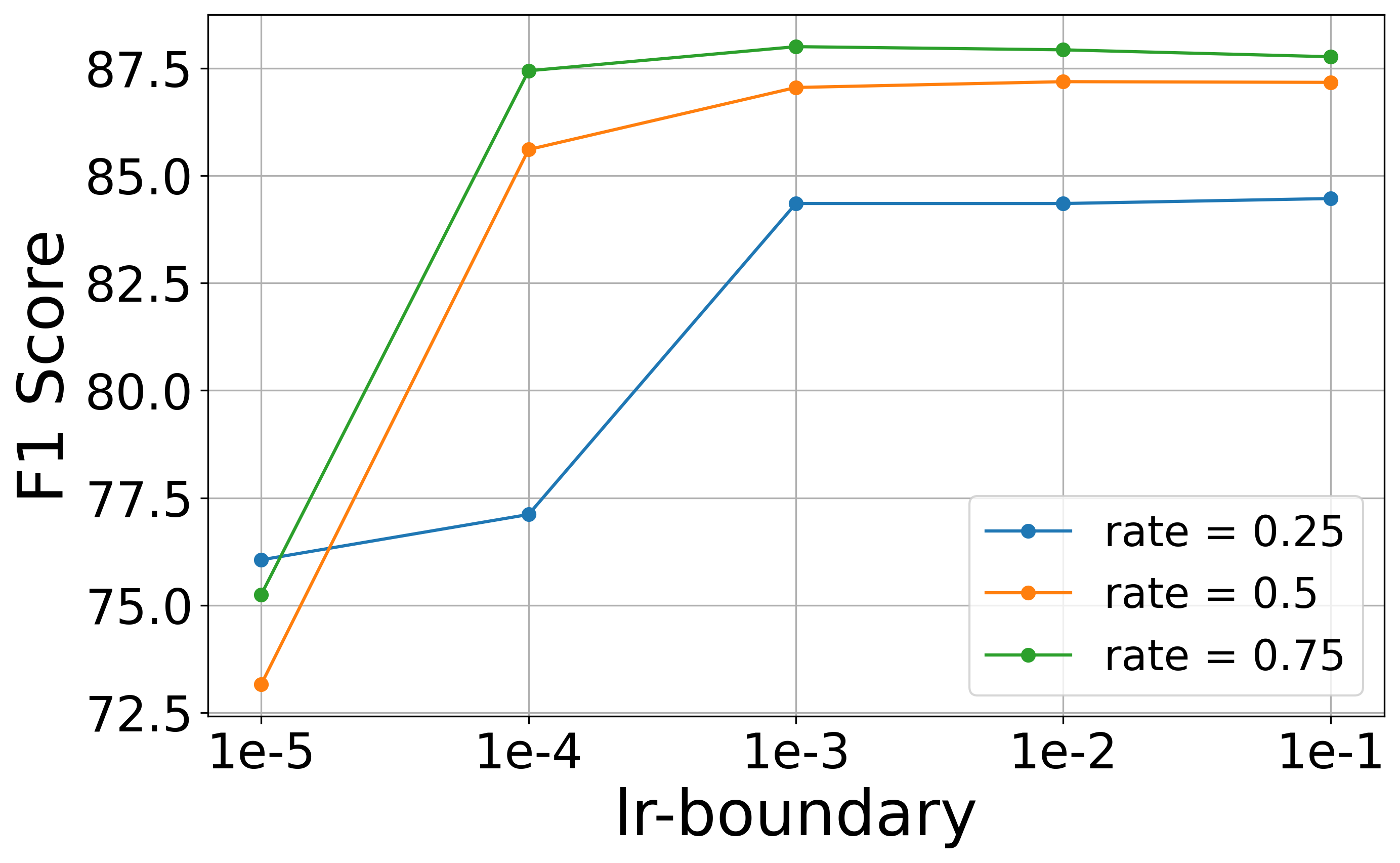}
        \caption{F1 for Learning Rate}
    \end{subfigure}
    \begin{subfigure}[b]{0.33\textwidth}
        \includegraphics[width=\textwidth]{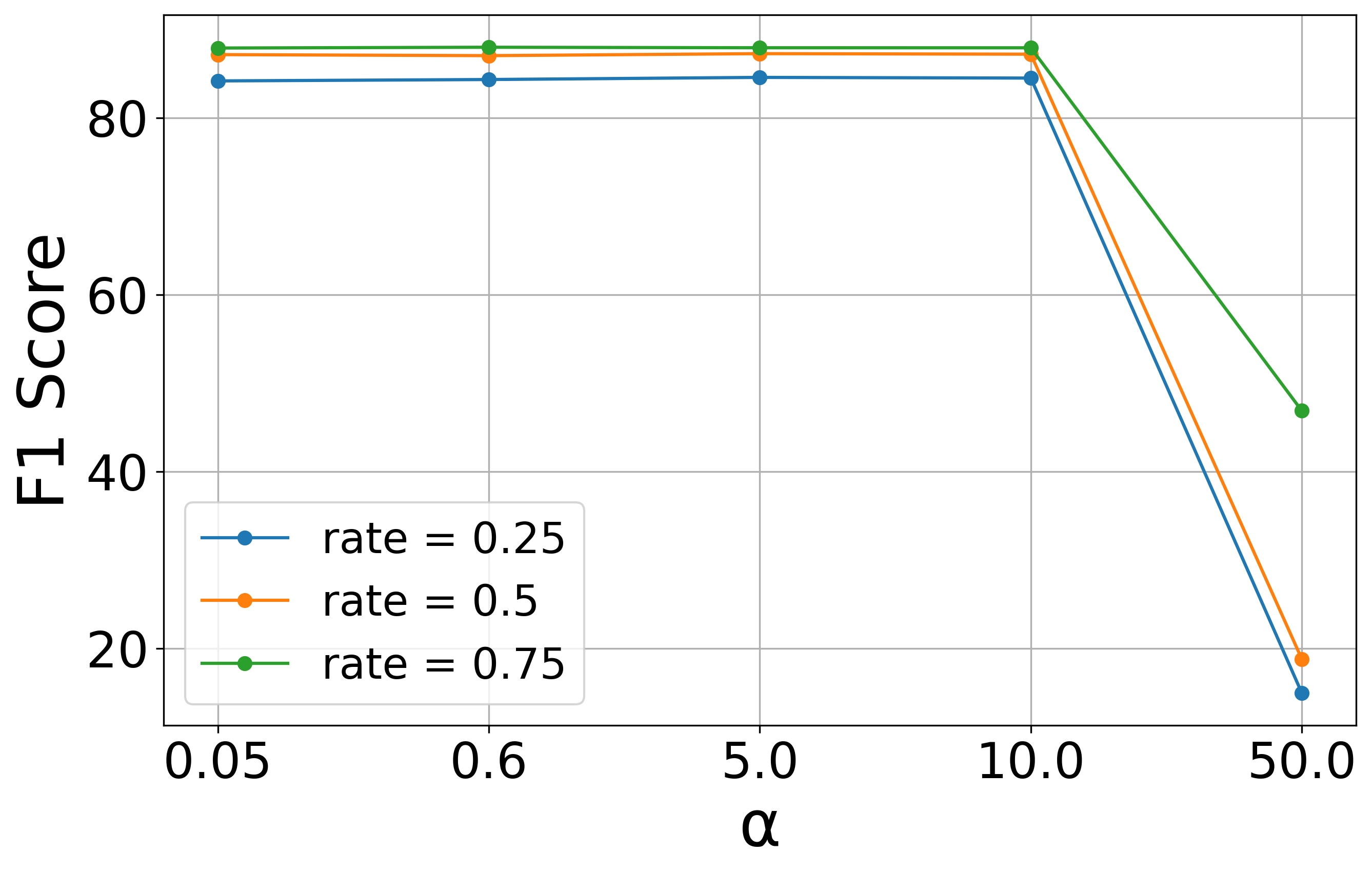}
        \caption{F1 for \(\alpha\)}
    \end{subfigure}
    \begin{subfigure}[b]{0.33\textwidth}
        \includegraphics[width=\textwidth]{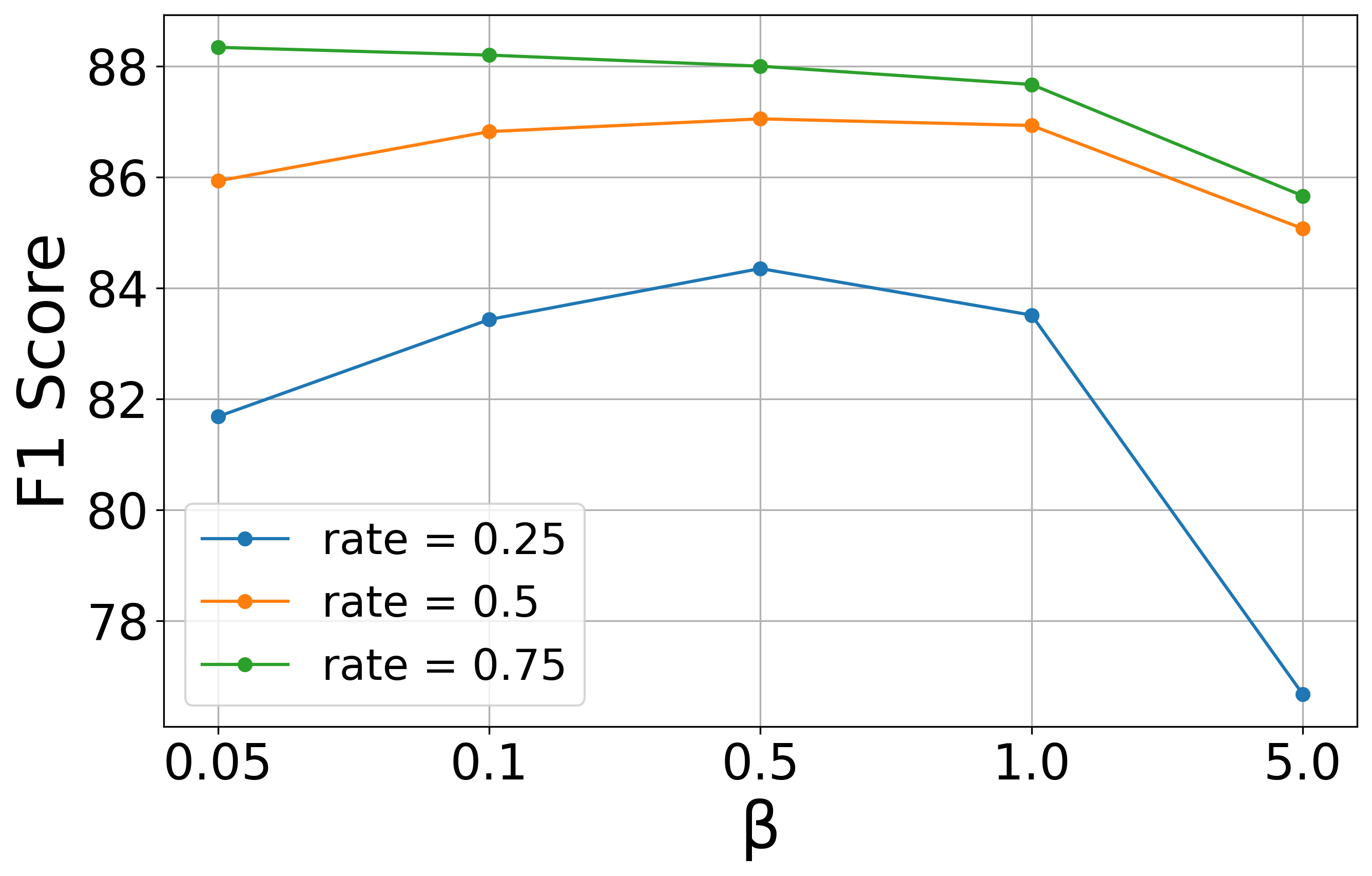}
        \caption{F1 for \(\beta\)}
    \end{subfigure}
    
    \begin{subfigure}[b]{0.33\textwidth}
        \includegraphics[width=\textwidth]{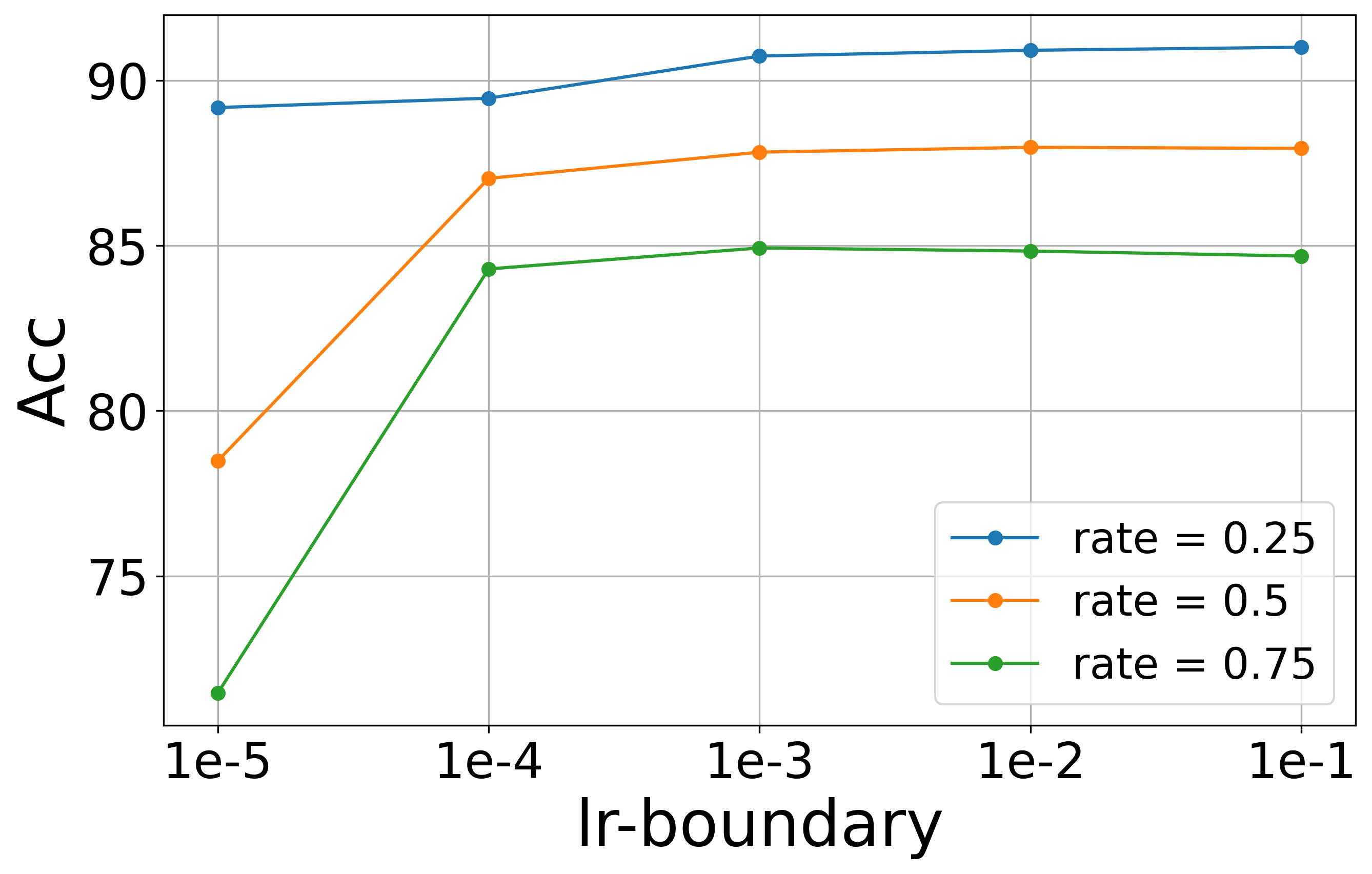}
        \caption{Acc for Learning Rate}
    \end{subfigure}
    \begin{subfigure}[b]{0.33\textwidth}
        \includegraphics[width=\textwidth]{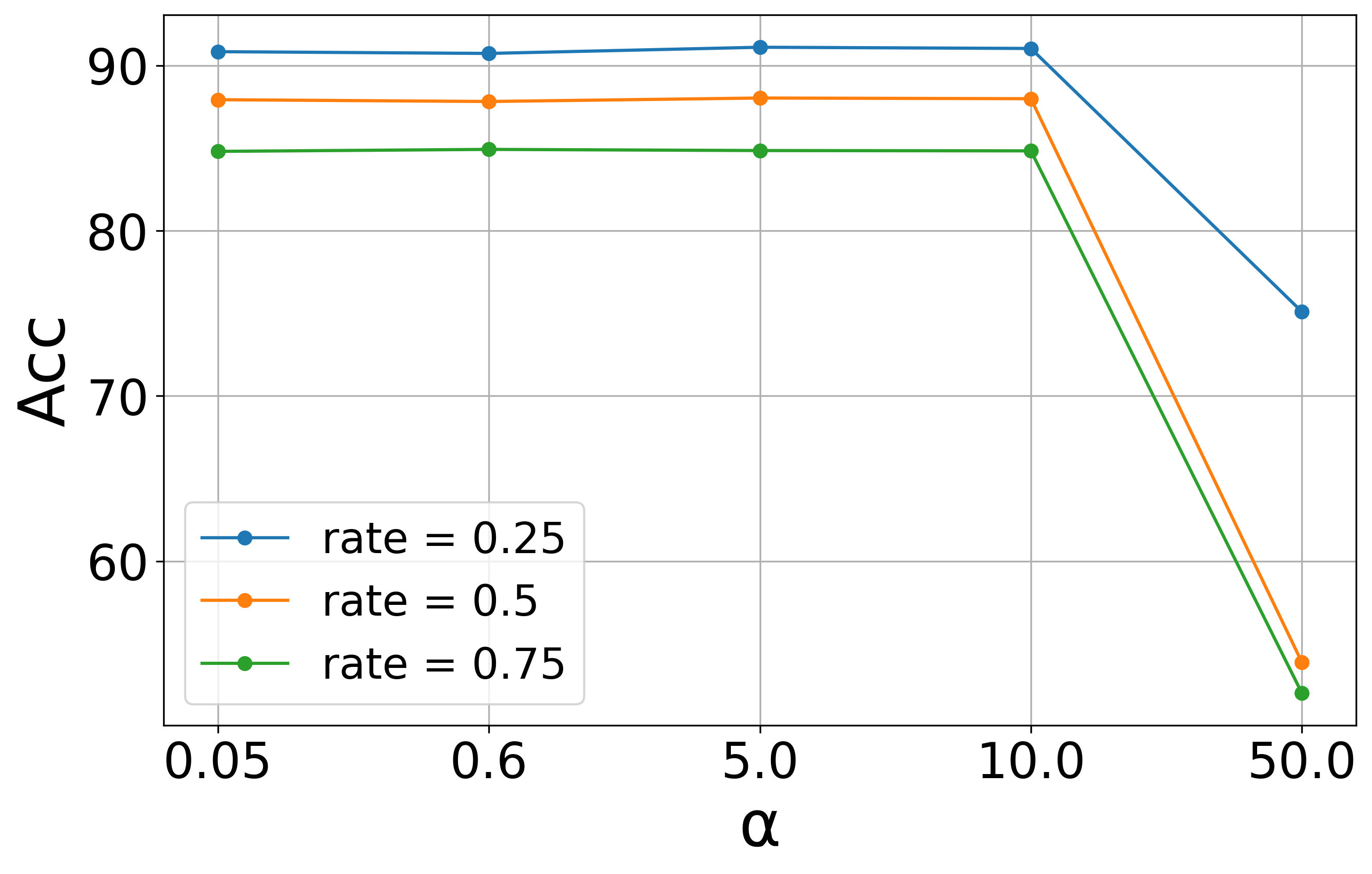}
        \caption{Acc for \(\alpha\)}
    \end{subfigure}
    \begin{subfigure}[b]{0.33\textwidth}
        \includegraphics[width=\textwidth]{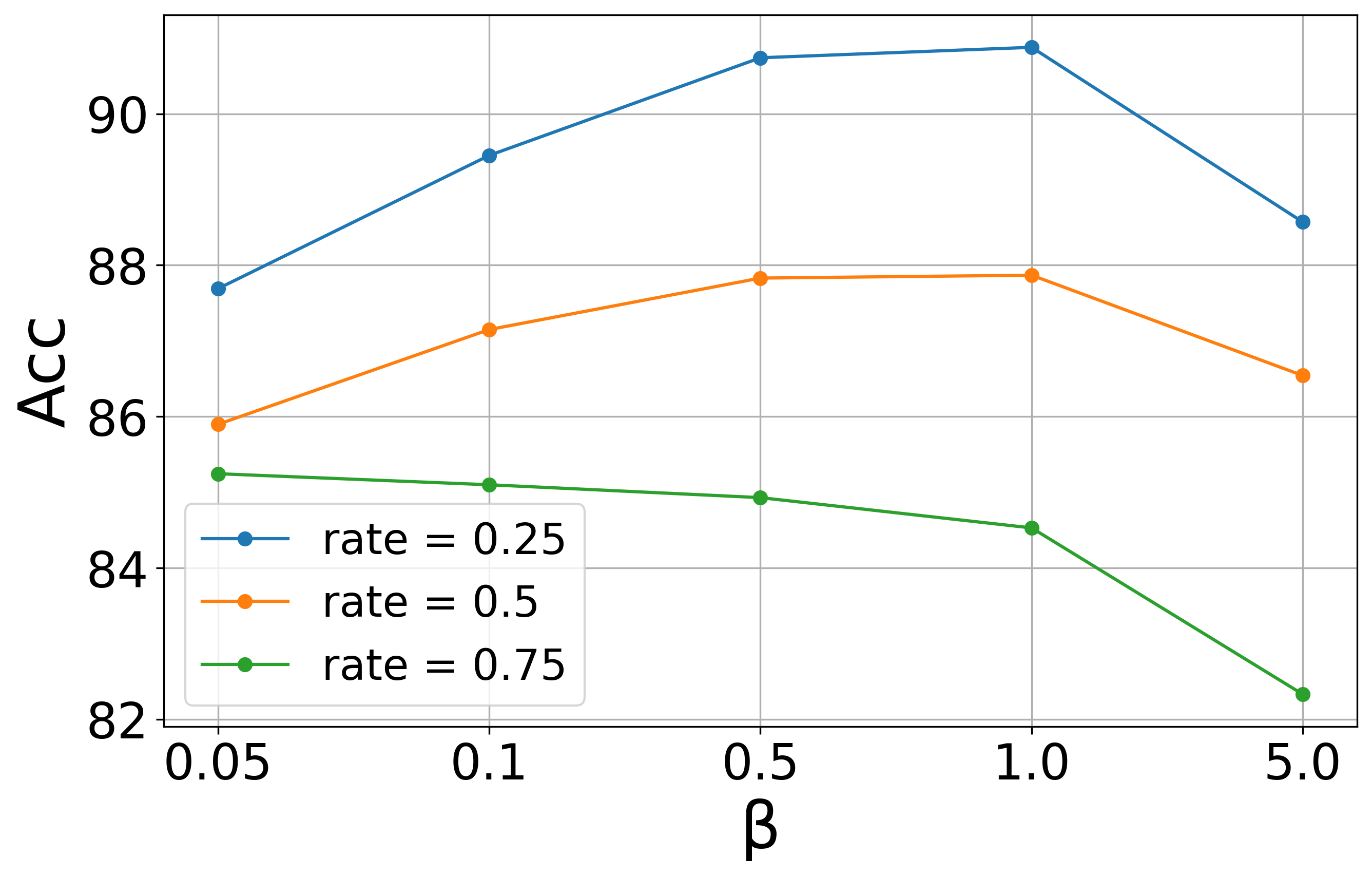}
        \caption{Acc for \(\beta\)}
    \end{subfigure}
    
    \caption{Comparison of F1 and Accuracy for different hyperparameters on StackOverflow.}
    \label{fig:all_images_stackoverflow}
\end{figure*}

To investigate the impact of hyperparameters on the experimental results, we conducted experiments focusing on three hyperparameters: the learning rate of boundary optimization $\text{lr}$, the parameter \(\alpha\) of Dirichlet distribution used in negative sample generation, and the penalty strength. The results on two datasets OOS and StackOverflow are presented in Figure~\ref{fig:all_images_oos} and Figure~\ref{fig:all_images_stackoverflow} respectively. 

$\text{lr}$ ranging from 1e-3 to 1e-1 yields high and stable F1 scores and accuracy. A low learning rate slows down the update process and leads to an insufficient boundary expansion, resulting in a significant drop of performance as the learning rate decreases. Conversely, a high learning rate does not provide noticeable performance advantages.

We observe that varying \(\alpha\) within a reasonable range (0.05 to 10) has a negligible impact on model performance. However, when \(\alpha\) exceeds a critical threshold between 10 and 50, the performance declines sharply on StackOverflow, as the negative samples concentrate in the gaps between known classes, inducing only marginal compression of decision boundaries.

As \(\beta\) increases, the performance initially improves before plateauing and eventually declining. Smaller \(\beta\) values produce insufficient boundary contraction, leaving the decision region overly expansive, resulting in false acceptance of open samples. Excessively large \(\beta\) induces an excessly strong penalty on the boundary causing false rejection of known samples.

\newpage
\bibliography{aaai2026}

\end{document}